\documentclass[lettersize,journal]{IEEEtran}
\usepackage{amsmath,amsfonts}
\usepackage{algorithmic}
\usepackage{algorithm}
\usepackage{subfigure}
\usepackage{array}
\usepackage[caption=false,font=footnotesize,labelfont=rm,textfont=rm]{subfig}
\usepackage{textcomp}
\usepackage{stfloats}
\usepackage{url}
\usepackage{verbatim}
\usepackage{graphicx}
\usepackage{subfig}
\usepackage{cite}
\usepackage{graphicx}
\usepackage{amsmath}
\usepackage{amssymb}
\usepackage{booktabs}
\usepackage{multirow}
\usepackage{bbding}
\usepackage{xcolor}
\usepackage{array}
\usepackage[numbers]{natbib}
\usepackage{threeparttable}

\definecolor{redcolor}{rgb}{0.761, 0.328, 0.263}
\definecolor{bluecolor}{rgb}{0.192, 0.545, 0.886}

\definecolor{darkred2}{rgb}{0.7, 0.0, 0.0}

\begin{document}
	
	\title{Beyond Overfitting: Doubly Adaptive Dropout for Generalizable AU Detection}
	
	\author{Yong Li,
		Yi Ren,
		Xuesong Niu,
		Yi Ding,
		Xiu-Shen Wei,
		Cuntai Guan ~\IEEEmembership{Fellow,~IEEE}
		\thanks{Corresponding author: Cuntai Guan.}
		\thanks{Yong Li is with the School of Computer Science and Engineering and the Key Laboratory of New Generation Artificial Intelligence Technology and Its Interdisciplinary Applications, Southeast University, Nanjing 210096, China. Email: mysee1989@gmail.com.}
			
		\thanks{Yi Ren are with the School of Computer Science and Engineering, Nanjing University of Science and Technology, Nanjing, 210094 China (e-mail: (mysee1989@gmail.com, yi.ren@njust.edu.cn)}
		\thanks{Xuesong Niu is with the Beijing Institute for General Artificial Intelligence, Beijing, China (e-mail:nxs583966@163.com).}
		\thanks{Xiu-Shen Wei is with School of Computer Science and Engineering, and Key Laboratory of New Generation Artificial Intelligence Technology and Its Interdisciplinary Applications, Southeast University,  Nanjing 210096, China (e-mail: weixs.gm@gmail.com).}
		\thanks{Cuntai Guan and Yi Ding are with the School of Computer Science and Engineering, Nanyang Technological University, 50 Nanyang Avenue, Singapore, 639798. E-mail: {(ctguan, ding.yi)}@ntu.edu.sg.}
		
		\thanks{Manuscript received April 19, 2021; revised August 16, 2021.}}
	
	\markboth{Journal of \LaTeX\ Class Files,~Vol.~14, No.~8, August~2021}%
	{Shell \MakeLowercase{\textit{et al.}}: A Sample Article Using IEEEtran.cls for IEEE Journals}
	
	
	\maketitle
	
\begin{abstract}
Facial Action Units (AUs) are essential for conveying psychological states and emotional expressions. While automatic AU detection systems leveraging deep learning have progressed, they often overfit to specific datasets and individual features, limiting their cross-domain applicability. To overcome these limitations, we propose a doubly adaptive dropout approach for cross-domain AU detection, which enhances the robustness of convolutional feature maps and spatial tokens against domain shifts. This approach includes a Channel Drop Unit (CD-Unit) and a Token Drop Unit (TD-Unit), which work together to reduce domain-specific noise at both the channel and token levels. The CD-Unit preserves domain-agnostic local patterns in feature maps, while the TD-Unit helps the model identify AU relationships generalizable across domains. An auxiliary domain classifier, integrated at each layer, guides the selective omission of domain-sensitive features. To prevent excessive feature dropout, a progressive training strategy is used, allowing for selective exclusion of sensitive features at any model layer. Our method consistently outperforms existing techniques in cross-domain AU detection, as demonstrated by extensive experimental evaluations. Visualizations of attention maps also highlight clear and meaningful patterns related to both individual and combined AUs, further validating the approach's effectiveness.
\end{abstract}
	
	\begin{IEEEkeywords}
		Facial Action Unit Detection, Domain Adaption, Feature Dropout
	\end{IEEEkeywords}
	
\section{Introduction}
\label{sec:intro}
	
		\begin{figure}[h]
		\centering
		\includegraphics[width=0.48\textwidth]{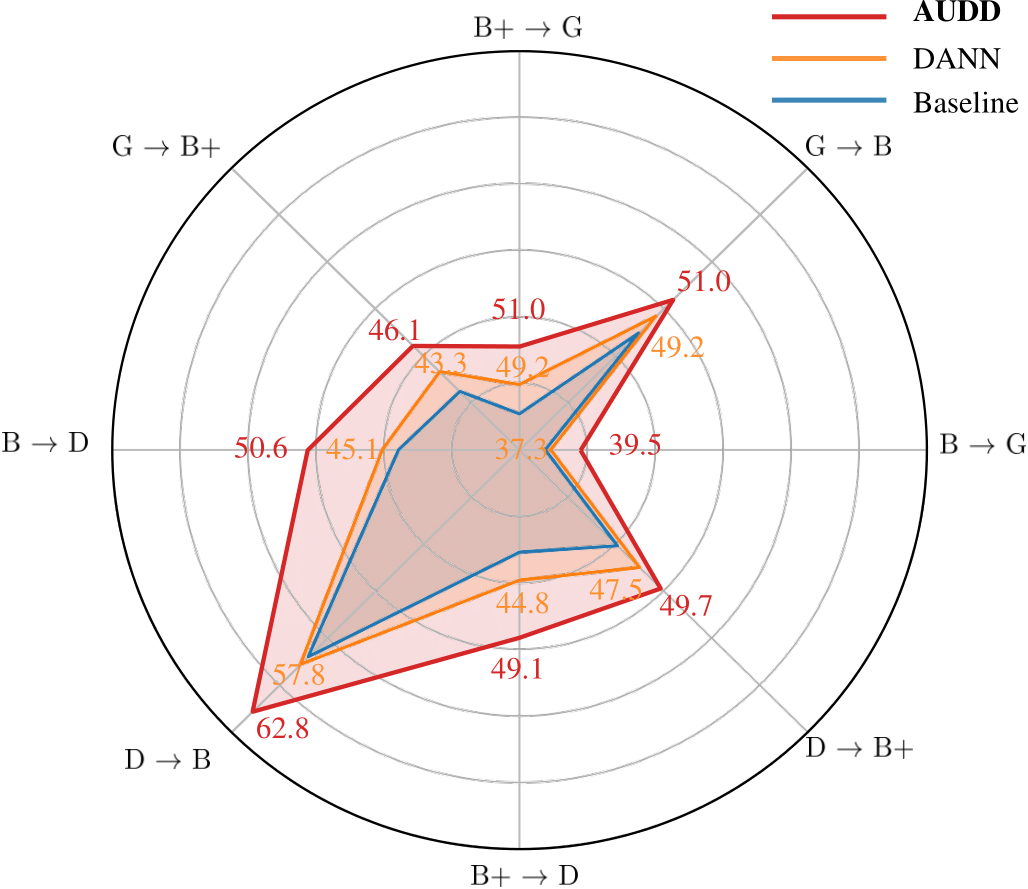}
		\caption{ Our proposed AUDD shows its superiority on a broad range of 8 cross-domain AU detection scenarios. Here \textit{B}, \textit{B+}, \textit{D}, \textit{G} denotes BP4D, BP4D+, DISFA, GFT dataset, respectively. Baseline denotes directly  evaluating the AU detection performance on the target dataset. More qualitative comparison can be found in Sec.~\ref{sec:exp_ablation_study}.
		}
		\label{fig:figure1}
	\end{figure}

Facial expressions constitute a nuanced channel for conveying a wide array of meanings, encompassing emotions, intentions, attitudes, and both mental and physical states~\cite{picard2000affective}. The Facial Action Coding System (FACS) \cite{friesen1978facial}, which is firmly rooted in anatomical principles, stands as the most exhaustive approach for encoding and representing facial expressions. FACS offers an explainable and reliable framework for the analysis of human facial expressions. It systematically dissects facial expressions into discrete elements termed Action Units (AUs), which correspond to the activation of specific facial muscles or muscle groups. By identifying and quantifying these AUs, researchers can methodically investigate facial expressions and the emotional states they signify. Presently, automated AU detection has emerged as a burgeoning field within computer vision, holding considerable potential for diverse applications such as human-computer interaction, emotion analysis, and medical pain assessment~\cite{li2018occlusion, lucey2009automatically, niu2023cnn}.

In recent times, deep neural networks (DNNs) have been increasingly utilized for within-dataset AU detection, yielding significant advancements~\cite{wang2017expression, li2021meta}. However, their performance tends to deteriorate noticeably when confronted with new scenarios or evaluated across different datasets~\cite{ghosh2015multi, onal2019cross, ertugrul2020crossing, yin2021self}. As demonstrated in recent work~\cite{yin2021self}, the AU feature extractor tends to produce indiscriminate features for unseen data from previously unseen domains.
Moreover, the process of annotating AUs is inherently challenging due to the subtle facial deformations induced by AUs, which span various local facial regions with varying intensities~\cite{chen2022geoconv}. Consequently, manually annotating and constructing AU detection models for each novel scenario entails considerable time and effort.
Given the potential disparities across domains and the intricacies associated with manual AU annotation, it becomes imperative to devise innovative methodologies. These methodologies should facilitate the adaptation of AU detection models to new scenarios without reliance on annotated datasets specific to these novel environments.


	

	
	
This manuscript mitigates the challenges posed by cross-domain AU detection through the introduction of a novel methodology termed Doubly adaptive Dropout (AUDD). AUDD is purposefully crafted to remove features that are susceptible to domain discrepancies while simultaneously preserving representations that are both invariant to domain variations and discriminative for AU detection. Traditionally, dropout techniques have been employed as a regularization strategy within fully-connected layers~\cite{srivastava2014dropout}, and subsequently adapted to enhance the regularization of convolutional features~\cite{hou2019weighted, ding2021channel, kong2022reflash}.
Despite the conceptual simplicity of discarding domain-sensitive features, the practical implementation of this strategy poses two significant challenges: (1) How to reliably and autonomously identify features that are sensitive to domain-specific variations? (2) How to develop a dropout-based framework that is meticulously tailored to enhance the efficacy of cross-domain AU detection? 

To respond the first challenge of autonomously identification of domain-sensitive features, AUDD incorporates a lightweight, trainable domain classifier. This binary classifier is adept at evaluating features derived from Convolutional Neural Networks (CNNs) or transformers to ascertain their relevance for domain classification. The efficacy of this classifier in accurately distinguishing between domains serves as a proxy for gauging the domain-specific sensitivity of the features. Addressing the second challenge, which involves adapting the feature-dropping framework for enhanced cross-domain AU detection, necessitates a sophisticated neural network design. Our approach leverages a hybrid architecture that synergizes an initial CNN layer with subsequent transformer layers, as detailed in~\cite{dosovitskiy2020image}. This strategic combination empowers our AUDD to concurrently mitigate domain-sensitive channels and tokens. Given that each spatial token correlates with a specific facial region reflective of certain AUs, the targeted elimination of tokens extends beyond mere domain-specific feature reduction. It also aids the model in in deducing the underlying connections that are consistent across datasets, enhancing the model’s generalisation capabilities. This dual-faceted approach not only can be used to drop the domain-sensitive features, but also used to assist the model to implicitly inferring the intrinsic relationships between AUs, thus enhancing cross-domain generalizability.

Specifically, our proposed AUDD consists of two integral components: (1) Channel Drop Unit (CD-Unit) and (2) Token Drop Unit (TD-Unit). These two components function collaboratively to mitigate domain-specific perturbations, channel-wise and token-wise, respectively. The CD-Unit is designed to autonomously identify and then drop the convolutional channels with undesired domain-sensitive attributes. Furthermore, the TD-Unit aims to perceive and preserve local tokens with generalizable features. With the TD-Unit, AUDD is expected to implicitly deduce the inherent AU relationships. It is anticipated to possess significant cross-dataset applicability because each token focuses on specific local areas of the input facial images~\cite{li2021meta}. Through the strategic dropping and retention of features, AUDD aspires to achieve a delicate balance between domain adaptability and the preservation of AU-specific information.

Furthermore, previous works have verified that the presence of domain-sensitive features actually exist across both shallow and deep layers of the network~\cite{yosinski2014transferable}, challenging the conventional dropout practices that predominantly target either high-level or low-level features. In response, we adopt a progressive training strategy wherein, during each iteration, a CNN block preceding and a transformer block succeeding are chosen at random, followed by the application of a CD-Unit and TD-Unit for adaptive feature dropout.
This progressive training strategy ensures a comprehensive attenuation of domain-sensitive channels and tokens throughout the network's hierarchy. Concurrently, it inherently safeguards against the excessive exclusion of features, a common pitfall that could precipitate training instability. 
As illustrated in Fig.~\ref{fig:figure1}, AUDD achieves superior cross-domain AU detection accuracy than DANN~\cite{ganin2015unsupervised} and the typical baseline methods under various settings. More experimental details and comparisons can be found in Sec.~\ref{sec:experiments}.

Our main contributions can be concluded as follows:
\begin{itemize}
	\item We introduce an innovative doubly adaptive feature selection approach aimed at improving cross-domain facial AU detection. The proposed AUDD method deliberately diminishes domain-specific features to ensure the adaptability of AU detection across various domains.
	\item  AUDD seamlessly integrates CD-Unit and TD-Unit, which collaboratively suppress domain-related nuisances while retaining features that are both generalizable and discriminative for AU detection. These units introduce a minimal number of trainable parameters and impose no additional computational load during the inference phase.
	\item Experimental results on various cross-dataset settings show that AUDD consistently improves facial AU detection performance in new scenarios. AUDD also outperforms other representative unsupervised domain adaptation AU detection approaches. Visualization results demonstrate meaningful distribution patterns regarding both individual and combined AUs, indicating AUDD's efficacy in capturing relevant facial dynamics for AU detection.
\end{itemize}

The remainder of this paper is organized as follows. In Section \ref{sec:related work} we review related work in facial action unit detection, domain adaptation and dropout. In Section \ref{sec:method}, we present the details of our AUDD method, while in Section \ref{sec:experiments}, we demonstrate its efficacy through experiments.

\section{Related Work}
\label{sec:related work}
\textbf{Facial action unit detection and unsupervised domain adaptation.} Automatic AU detection has been studied for decades and a number of approaches have been proposed.
Given that AUs correspond to subtle alterations in facial muscle regions and skin textures, numerous methods leverage facial landmarks ~\cite{ge2023algrnet} or employ adaptive attention mechanisms~\cite{shao2021jaa, wang2022action} to acquire region-specific representations.
Acknowledging the inherent dependencies and exclusions among different facial action units, some literature suggests explicitly incorporating structural information to enhance the robustness of AU detection. 
On the other hand, the limited availability of training data for AU detection necessitates exploration into alternative methodologies. Some studies delve into self-supervised learning approaches ~\cite{chang2022knowledge, li2019self} and weakly-supervised methods~\cite{peng2018weakly} to augment the generalizability of AU detectors.
Reference~\cite{li2020learning} introduced a self-supervised learning framework aimed at acquiring pose-invariant AU features by predicting separate displacements for pose and AU between randomly sampled facial frames belonging to the same subject. Similarly, Chang et al.~\cite{chang2022knowledge} proposed a contrastive learning component that capitalizes on inter-area differences to acquire AU-related local representations while preserving intra-area instance discrimination. However, existing AU detection models usually exhibit significant overfitting issue on the training data, resulting in reduced performance when confronted with out-of-distribution or cross-dataset data~\cite{onal2019cross, ertugrul2020crossing, yin2021self}. Such obvious AU detection accuracy degradation caused by data distribution discrepancy hinders the applications of current AU detectors, as in reality the various new scenarios usually exhibit different data distributions. 

In this manuscript, we conduct a comparative analysis of the proposed method with typical Unsupervised Domain Adaptation (UDA) techniques, including Domain-Adversarial Neural Network (DANN)~\cite{ganin2015unsupervised}, MCD~\cite{saito2018maximum}.
Ganin et al.~\cite{ganin2015unsupervised} introduced DANN, a method employing an elegant domain classifier that reverses the gradient by multiplying it by a negative scalar during backpropagation to achieve unsupervised domain adaptation. 
Saito et al.~\cite{saito2018maximum} proposed Maximum Classifier Discrepancy (MCD), which aims to align the distributions of source and target domains by leveraging task-specific decision boundaries. 
While these approaches endeavor to align global source and target data distributions to learn domain-invariant representations, the resulting representations do not necessarily ensure the desired class-to-class alignment. Our proposed AUDD bypass this obvious drawback via simultaneously dropping the domain-sensitive features and  preserving representations that are discriminative for AU detection.

Furthermore, a recent trend in UDA methods focuses on feature-disentangling-based approaches, e.g., Lee et al.~\cite{lee2021dranet} proposed a method to disentangle image representations and transfer visual attributes in a latent space for UDA.  Chen et al.~\cite{chen2021joint}  introduced a joint generative and contrastive learning framework for unsupervised person re-identification.
Different with these feature-disentangling-based methods, AUDD does not rely on pixel-wise image reconstruction that is quite challenging and can mitigate the domain-sensitive nuisances more flexibly.
We will compare AUDD with the general UDA methods and present the results as well as the corresponding analysis in Sec.~\ref{sec:exp_sota}. 


\textbf{Dropout.}
Dropout has emerged as a potent algorithm for training robust deep networks, primarily due to its efficacy in mitigating overfitting by discouraging the co-adaptation of feature detectors~\cite{srivastava2014dropout, niu2023cnn}. Conceptually, Dropout can be viewed as an approximate model aggregation technique, wherein an exponential number of smaller networks are averaged to form a more robust ensemble. 

Based on the vanilla dropout, a plethora of variants have been introduced in the literature, e.g., Morerio et al.~\cite{morerio2017curriculum} proposed Curriculum Dropout, which challenges fixed dropout probabilities and employs a time scheduling mechanism to regulate the retention probability of neurons in the network.
To alleviate the overfitting issue when finetuning CNN on small datasets, Hou et al.~\cite{hou2019weighted} proposed Weighted Channel Dropout, a technique specifically designed for the regularization of convolutional layers. 
Subsequently, structured dropout methods have also garnered attention. 
Ghiasi et al.~\cite{ghiasi2018dropblock} introduced DropBlock, a structured dropout variant where units in a contiguous region of a convolutional feature map are dropped together. 
Lee et al.~\cite{lee2019drop} proposed Drop to Adapt (DTA), which utilizes adversarial dropout to learn discriminative features by enforcing the cluster assumption. 
However, none of the above approaches consider to drop feature from hybrid granularities, a critical design aspect incorporated in AUDD. AUDD uniquely integrates channel and token-wise dropout mechanisms for cross-domain action unit (AU) detection, distinguishing it from other methods.

Subsequently, Zeng et al.~\cite{zeng2020corrdrop} devised Correlation based Dropout (CorrDrop), a structural dropout method that regularizes CNNs by dropping feature units based on feature correlation. 
Guo et al.~\cite{guo2023domaindrop} presented a dropout-based domain generalization framework, aiming to enhance channel robustness to domain shifts.
In comparison to existing dropout-based domain adaptation or domain generalization methods, our proposed AUDD offers a unique advantage by being AU-specific. This specificity allows AUDD to provide more effective regularization effects, specifically targeting and mitigating the severe distribution discrepancy issues between the source and target domains.

\begin{figure*}[htb]
	\centering
	\includegraphics[width=1\textwidth]{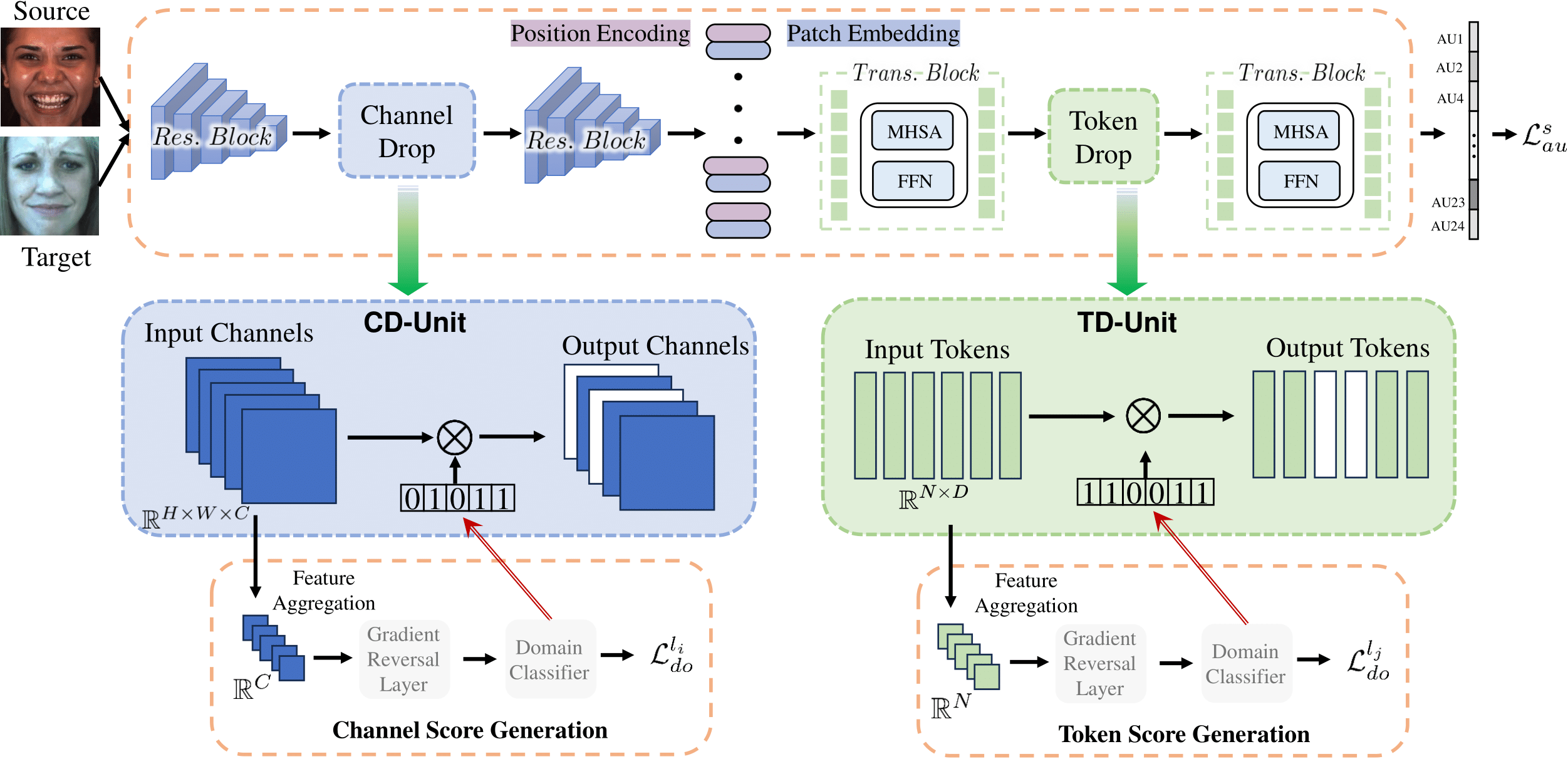}
	\caption{\textbf{(a)} The framework of the proposed AUDD for cross-domain AU detection. AUDD takes as input a hybrid batch of source and target images and encode the images with three preceding CNN-based residual blocks and three subsequent transformer blocks (Sec.~\ref{sec:hybrid_cnn_trans}). To automatically maximize the preservation of domain-agnostic and generalizable channels/tokens, AUDD employs both the CD-Unit and TD-Unit concurrently for adaptive feature weight generation and dropping (Sec.~\ref{sec:mask_gene}).Each channel/token is assigned a corresponding score to determine whether it is domain-sensitive (Sec.~III.C).
	}
	\label{fig:framework}
\end{figure*}

\section{Method}
\label{sec:method}

In this paper, we study the problem of cross-domain AU detection. Without loss of generality, suppose we have a source AU dataset $\mathcal{D}^{s} = \{(\mathcal{X}^s_{i}, y_{i}^j), 1 \leq i \leq N\}$ and a target AU dataset $\mathcal{D}^{t} = \{\mathcal{X}^t_{i}, 1 \leq i \leq M\}$.
Among them, $\mathcal{X}^s_i, \mathcal{X}^t_i$ denote the $i$-th image in $\mathcal{D}^{s}$, $\mathcal{D}^{t}$, respectively. $N, M$ denote the total number of samples in $\mathcal{D}^{s}$ and $\mathcal{D}^{t}$.
$y^j_{i} \in \{0, 1\}$ represents annotation w.r.t the $j$-th AU for $i$-th sample in $\mathcal{D}^{s}$. 1 means the AU is active, 0 means inactive. Under conventional UDA setting, merely the images in the source dataset $\mathcal{D}^{s}$ are annotated with $y_{i}$. Our target is to maximally align the feature distributions of  $\mathcal{D}^{s}$ and $\mathcal{D}^{t}$ without the access to the annotations in the target domain. We will detail the framework and training process bellow.

\subsection{Framework Overview}
We propose a Doubly Adaptive Dropout method, i.e., \texttt{AUDD}, to empower the simultaneous dropout of domain-sensitive convolutional channels and spatial tokens, thus fostering the retention of features with high generalizability across diverse granularities. The architecture of the AUDD is shown in Fig.~\ref{fig:framework}. 

The network takes facial images from both source and target datasets as input. These images are then passed through several CNN-based residual blocks, resulting in the generation of feature maps. For these generated feature maps, AUDD employs the CD-Unit to generate corresponding binary masks for automatically dropping the domain-sensitive channels. Subsequently, the convolutional feature maps from the last residual block are reshaped to generate input tokens for subsequent transformer blocks. As depicted in Fig.~\ref{fig:framework}, AUDD simultaneously employs the TD-Unit to adaptively drop and filter the intermediate tokens. Regarding the CD- and TD-Units, the former is randomly interspersed among the residual blocks, and similarly, the TD-Unit is incorporated randomly within the transformer part. These units operate by autonomously generating a binary mask that discerns the significance of respective channels and tokens in the context of cross-domain AU detection.

For the facial images from the source domain, we exploit the multi-label sigmoid cross-entropy loss for AU detection. It is formulated as
\begin{equation}
	\mathcal{L}_{AU} = - \sum^{J}_{j} y^j \log\hat{y}^{j} + (1 - y^{j}) \log (1-\hat{y}^{j})
\end{equation}
where $J$ is the number of AUs. $y^j$ denotes the $j$-th ground truth AU label of the input AU sample. $\hat{y}^j$ means the predicted AU score.  Below, we will present the detailed neural network structure in AUDD in Sec.~\ref{sec:hybrid_cnn_trans}, how AUDD learns to drop domain-sensitive features with CD-/TD-Units in Sec.~\ref{sec:mask_gene}.

\subsection{Hybrid CNN-Transformer Architecture}
\label{sec:hybrid_cnn_trans}
To fully leverage the dropout mechanism for suppressing the domain-specific features while concurrently preserving the generalizable attributes across various granularity,  AUDD employs a hybrid neural network architecture, combining Convolutional Neural Networks (CNNs) and Transformer blocks, akin to the structure presented in~\cite{dosovitskiy2020image}. Through the integration of a CNN-based residual block, AUDD is adept at channel-wise identification and elimination of domain-specific features, facilitated by the subsequent CD-Unit, wherein each channel potentially encapsulates distinct semantic patterns. In parallel, the transformer-based blocks of the network are used to preserve the generalizable features from an alternative viewpoint, with each token representing localized facial regions. This dual approach enables AUDD  to drop the undesired features and deduce the inherent relationships among AUs to enhance its generalizability. The details of the hybrid architecture are as follows. We also show the parameters configuration in Tab.~\ref{Tab:strc_table}.

\begin{itemize}
	\item[(1)] The CNN component comprises four successive residual stages, akin to ResNet-50. These stages are utilized to extract detailed spatial features from the input facial image.
	Following the methodology outlined in reference~\cite{dosovitskiy2020image}, we replaced the fourth stage of ResNet-50 with an equal number of layers as in stage 3. So the total number of layers are unchanged and we can obtain \texttt{three} residual blocks.
	The output feature maps $\mathbf{X}_{i}\in \mathbb{R}^{ H_{i} \times W_{i} \times C_i}$ for the $i$-th residual block are then flattened into $H_{i} \times W_{i}$ tokens, where each token has a $C_i$-dimensional feature vector. Typically, based on the $224 \times 224 \times 3$ input image, the feature maps $\mathbf{X}_{i\in{1,2,3}}$ for the three residual blocks are encoded as $56 \times 56 \times 256$, $28 \times 28 \times 512$, and $14 \times 14 \times 1024$, respectively. The final feature maps from the last residual block are then convolved with a pointwise convolution, resulting in reshaped feature maps $\mathbf{X}_r \in \mathbb{R}^{14 \times 14 \times 768}$.
	\item[(2)] The transformer part is based on the vanilla ViT-B/16 architecture from~\cite{dosovitskiy2020image}. It comprises 12 transformer encoders, each consisting of alternating layers of multiheaded self-attention (MHSA) and MLP layers. Analogous to the preceding CNN part, the total 12 transformer encoders are evenly partitioned into \texttt{three} successive blocks. The reshaped feature maps $\mathbf{X}_{r} \in \mathbb{R}^{14 \times 14 \times 768}$ are then flattened to obtain the input feature vector $\mathbf{x}_j \in \mathbb{R}^{196 \times 768}$, where $j$ denotes the index of the transformer block. This means each of the $196$ tokens corresponds to one spatial location in the vanilla convolutional feature map, representing a local facial region with respect to the input image. Here, the learnable class token is intentionally omitted to prevent a scenario where the exclusion of other tokens forces the class token to encapsulate all semantic information, rendering the remaining vanilla input tokens redundant. Subsequently, each token is augmented with a 1-D position embedding and fed into the transformer blocks to model spatial relationships for AU detection.
	
\end{itemize}

\begin{table}[ht]
	\caption{Architectures for AUDD. Building blocks are shown in brackets with the numbers of blocks stacked.}
	\centering
	\scalebox{0.95} {\begin{tabular}{c|c|c}
			\toprule
			
			Block  & Operations & Output Size \\
			\midrule
			& $7 \times 7$, 64; $3 \times 3$ max pool, stride 2 & $56 \times 56 \times 64$ \\ \hline\hline
			1& [$1 \times 1$, 64; $3 \times 3$, 64; $1 \times 1$, 256] $\times 3$ & $56 \times 56 \times 256$ \\  \hline
			2& [$1 \times 1$, 128; $3 \times 3$, 128; $1 \times 1$, 512] $\times 4$ & $28 \times 28 \times 512$ \\  \hline
			3& [$1 \times 1$, 256; $3 \times 3$, 256; $1 \times 1$, 1024] $\times 9$ & $14 \times 14 \times 1024$ \\
			\hline \hline
			& $1 \times 1$, 768; Flatten; Position Embedding & $196 \times 768$\\ \hline\hline
			4& [dim 768; head 12] $\times 4$ & $196 \times 768$ \\ \hline
			5& [dim 768; head 12] $\times 4$ & $196 \times 768$\\ \hline
			6& [dim 768; head 12] $\times 4$ & $196 \times 768$\\
			\bottomrule
	\end{tabular}}
	\label{Tab:strc_table}
\end{table}

\subsection{Selective Feature Dropout}
\label{sec:mask_gene}

With the hybrid neural network architecture, we are capable of dropping the undesired channels or tokens if they are domain-sensitive. As illustrate in Fig.~\ref{fig:framework}, we insert a CD-Unit between two consecutive residual blocks to automatically filter the channels. In parallel, a TD-Unit can be placed between two consecutive transformer blocks to merely preserve the generalizable tokens.  Both CD-Unit and TD-Unit function in two steps: (1) Rate the channels or tokens according to their contribution to the true domain prediction; (2) Adaptively drop the channels or tokens according to both the corresponding score and a Weighted Random Selection algorithm~\cite{efraimidis2006weighted}.

\textbf{Domain-sensitive score generation.} We first calculate the domain-sensitive score to determine which channels or tokens are domain-specific and need to be dropped. The scores are calculated based on the parameters of the domain discriminator $F_{d}$, which is used for domain classification and consists of a Global Average Pooling (GAP) layer $F_{GAP}$, a Gradient Reversal Layer (GRL) $F_{GRL}$, and a Fully-Connected (FC) layer $F_{FC}$. The GAP layer is used to obtain global information. For the CNN part, the feature of the $i$-th CNN block $\mathbf{X}{i}\in \mathbb{R}^{H{i} \times W_{i} \times C_{i}}$ is pooled into a global feature $\mathbf{X}{gi} \in \mathbb{R}^{C{i}}$. For the transformer part, the global token of the $j$-th block $\textbf{x}{gj} \in \mathbb{R}^{N{j}}$ is pooled from its input tokens $\textbf{x}{gj}\in \mathbb{R}^{N{j}\times D_{j}}$, where $N_{j}$ tokens are of $D_{j}$ dimension. Following~\cite{ganin2015unsupervised}, to avoid the negative impact of domain discriminators on the main network, we insert a GRL layer to truncate the gradients before the FC layer. The loss for domain classification is formulated as:
\begin{equation}
	\mathcal{L}_{do}^{l_{i}} = -\frac{1}{2}(\frac{1}{M}\sum_{i}^{M} \log \hat{p}^s_{i} + \frac{1}{N}\sum_{i}^{N} \log \hat{p}^t_{i})
\end{equation}
where $\hat{p}^s_{i}$, $\hat{p}^t_{i}$ demote the predicted probability for the input feature, i.e., $\mathbf{X}_{i}$ or $\mathbf{x}_{j}$, belonging to the source or target domain, $l_i$ corresponds to the $i$-th residual or transformer block in AUDD.

We then calculate the domain-sensitive scores based on the domain discriminator. We assume that the more features contribute to true domain classification, the more these features are likely to contain domain-specific information. Therefore, we calculate the correlation between the input channels/tokens and domain-specific information using the weighted activations of the domain classification. To be specific, for the generated features $f_{input}^{d} \in \{\mathbf{X}_{i}^{d}, \mathbf{x}_{j}^{d}, 1 \leq i \leq 3, 1 \leq j \leq 3\}$ of different CNN and transformer blocks from domain $d \in \{s, t\}$, we formulate them as a set of $K$ sub-features $\{f_{sub1}^{d}, f_{sub2}^{d}, \cdots, f_{subK}^{d}\}$ to be decided whether should be dropped, where $K$ is the size of the sub feature sets, i.e., the number of channels for the CNN part and the number of tokens for the transformer part. The domain-sensitive score for the $j$-the sub-feature of input feature $f_{input}^{d}$ is computed as 
\begin{equation}
	s_{j} = 
	\begin{cases} 
		W_{j}^{0} \cdot F_{GAP}(f_{subj}^{d}) & \text{if } d = s, \\
		W_{j}^{1} \cdot F_{GAP}(f_{subj}^{d}) & \text{if } d = t.
	\end{cases}
\end{equation}
where $W_{j}^{0}$ and $W_{j}^{1}$ is the layer weight of the $F_{FC}$ of domain discriminator for the source and target domain respectively. 

\textbf{Domain-irrelated feature selection.} With the domain-sensitive score, we can then calculate which channels or tokens should be dropped. Higher domain-sensitive score $s_i$ demonstrates the higher correlation between the sub-feature and domain-specific information. Therefore, we should drop these sub features with higher domain-sensitive scores. 
For the computing efficiency, we follow~\cite{hou2019weighted} and utilize Weighted Random Selection (WRS) to generate the binary mask for dropout. Specifically, for the $i$-th channel with $s_i$ , we first randomly sample a number $r \in (0, 1)$ and compute a key value $k_i = r^{1/s_i}$. The $N_{mask}$ sub-features with the largest $k_i$ will be masked to 0, and the domain-irrelated feature selection mask is formulated as
\begin{equation}
	M_i = 
	\begin{cases} 
		0 & \text{if } i \in argmax(\{k_1, k_2, \cdots, k_K\}, N_{mask}), \\
		1 & otherwise.
	\end{cases}
\end{equation}
where $argmax(\{k_1, k_2, \cdots, k_K\}, N_{mask})$ means the index of top $N_{mask}$ values in the list ${k_1, k_2, \cdots, k_K}$. The generated mask is then used as the dropout mask for further training. 

\begin{algorithm}
	\caption{Training Pipeline for AUDD}
	
	\textbf{Input:\space}Source images $\mathcal{X}^s$, target images $\mathcal{X}^{t}$, source labels $y^{s}$\\
	\textbf{Output:\space} Optimized model parameters $\theta$ \\
	\begin{algorithmic}[1]
		\FOR{$\mathcal{X}_i$ in $\mathcal{X}^s \bigcup \mathcal{X}^{t}$}
		\STATE Generate intermediate CNN and transformer features $\{X_{i}, x_{j}, 1 \leq i \leq 3, 1 \leq j \leq 3\}$ for $\mathcal{X}_i$
		\STATE Initialize overall loss $L \leftarrow 0$ 
		\STATE Random sample one CNN feature $X_{i}$ from $\{X_{1}, X_{2}, X_{3}\}$ and one transformer feature $x_{j}$ from $\{x_{1}, x_{2}, x_{3}\}$ 
		\FOR{$f_{input}$ in $\{X_{i}, x_{j}\}$}
		\STATE Calculate $L_{do}^{l}$ and update $L \leftarrow L + L_{do}^{l}$
		\STATE Calculate domain sensitive score $s_i$ for $f_{input}$
		\STATE Generate a mask based on $s_i$
		\STATE Dropout sub-features of $f_{input}$ with mask
		\ENDFOR
		\IF{$\mathcal{X}_i\in \mathcal{X}^s$}
		\STATE Calculate $L_{AU}$ and update $L \leftarrow L + L_{AU}$
		\ENDIF
		\STATE Optimize $\theta$ with $L$
		\ENDFOR 
	\end{algorithmic}
	\label{alg:training}
\end{algorithm}

\textbf{Overall Objective.} 
To avoid drop too much features that would case training instabilities, we propose a progressive training strategy that randomly insert a CD-Unit and a TD-Unit w.r.t the three residual and transformer blocks, respectively. Let $\mathcal{L}^{l_i}_{do}$ and $\mathcal{L}^{l_j}_{do}$ respectively represent the domain classification loss generated by the CD-Unit and the TD-Unit,  the full objective function of our proposed AUDD can be formulated as,
\begin{equation}
	\mathcal{L}_{total} =  \mathcal{L}_{AU} + \alpha  \mathcal{L}^{l_i}_{do} + \beta \mathcal{L}^{l_j}_{do}
\end{equation}
where $\alpha$ and $\beta$ means the trade-off factors that control the importance of the domain classification constraint. For binary domain classification, we exploit the GRL operation, i.e., reverse the gradient from the domain classifier by a  constant value $\lambda$, which will be investigated in Sec.~\ref{sec:exp_ablation_study}.

The overall training process is illustrated in Alg.~\ref{alg:training}.
	
\section{Experiments}
\label{sec:experiments}

\subsection{Experiment Setup}
\textbf{Data Preparation.}
We evaluate the proposed AUDD on four publicly FACS  datasets, including BP4D~\cite{zhang2014bp4d}, BP4D+~\cite{zhang2016multimodal}, DISFA~\cite{mavadati2013disfa}, GFT~\cite{girard2017sayette}. \textbf{BP4D} encompasses data from 41 individuals (23 females and 18 males), yielding approximately 146,000 image frames annotated with 12 distinct AUs.  \textbf{BP4D+} comprises 140 subjects with a corpus of roughly 198,000 frames, each annotated with the same 12 AUs as BP4D. For the purpose of within-domain assessment, the two datasets were partitioned into three segments based on participant identifiers, facilitating a tripartite cross-validation scheme. The \textbf{DISFA} dataset, featuring 27 participants, includes close to 130,000 frames, each labeled with 8 AUs and corresponding intensity levels ranging from 0 to 5. Frames exhibiting intensities above the threshold of 1 were designated as positive instances, while the remainder were classified as negative. A subject-independent three-fold cross-validation protocol was employed for evaluation. Lastly, the \textbf{GFT}  dataset, documenting naturalistic social interactions among 96 participants divided into 32 triads, presents a unique challenge for AU detection due to the moderate degree of out-of-plane head movements. Each participant in the GFT dataset is associated with approximately 1,800 frames, annotated with 10 AUs.

\begin{table*}[ht]
	\centering
	\caption{F1-scores of domain adaptation between BP4D and GFT. \textbf{Bold} means the best.}
	\label{tab:4_bp4d_gft}
	\scalebox{0.95} {\begin{tabular}{c|c|c|c|c|c|c|c|c|c|c|c}
			\toprule
			AU & 1 & 2 & 4 & 6 & 10 & 12 & 14 & 15 & 23 & 24 & AVE\\
			\toprule
			\multicolumn{12}{c}{Source: \textbf{BP4D}. Target: \textbf{GFT}. } \\
			\midrule
			\textit{Source} & \textit{12.9} & \textit{28.7} & \textit{18.2} & \textit{60.8} & \textit{58.0} & \textit{69.2} & \textit{6.6} & \textit{24.8} & \textit{47.1} & \textit{42.4} & \textit{36.9}\\
			\midrule
			DANN~\cite{ganin2015unsupervised} & 5.9 & 17.4 & 2.8 & 51.4 & 38.3 & 52.4 & 6.0 & 18.1 & 36.1 & 20.9 & 24.9\\
			
			MCD~\cite{saito2018maximum} & 4.6 & 16.5 & 8.9 & 41.7 & 40.5 & 58.5 & 6.1 & 21.8 & 36.3 & 22.7 & 25.8\\
			
			P-MCD~\cite{yin2021self} & 9.6 & 22.1 & 8.1 & 54.0 & 53.6 & 57.2 & \textbf{7.7} & 14.4 & 37.0 & 31.8 & 29.6\\
			\midrule
			TCAE ~\cite{li2019self} & 17.4 & 19.4 & 9.5 & 62.0 & 54.6 & 59.3 & 7.1 & 10.4 & 37.3 & 36.7 & 31.4\\
			DGNet++~\cite{zou2020joint} & 6.3 & 16.8 & 7.7 & \textbf{65.3} & 55.3 & 60.5 & 5.2 & 12.9 & 34.7 & 37.0 & 30.2\\
			
			DRANet~\cite{lee2021dranet} & 13.0 & 18.4 & 8.4 & 61.3 & 56.1 & 55.8 & 7.3 & 19.7 & 32.8 & 40.3 & 31.3\\
			
			GCL~\cite{chen2021joint} & 14.2 & 4.3 & 7.7 & 61.4 & 52.4 & 61.1 & 6.1 & 23.4 & 24.2 & 44.5 & 29.9\\
			
			DTA~\cite{lee2019drop} & 8.6 & 21.8 & 6.6 & 43.4 & 39.6 & 45.1 & 0.0 & 17.4 & 20.0 & 24.7 & 22.7\\
			
			\textbf{AUDD(Ours)} & \textbf{17.8} & \textbf{29.6} & \textbf{18.5} & 60.4 & \textbf{64.7} & \textbf{70.9} & 7.4 & \textbf{26.5} & \textbf{52.1} & \textbf{47.0} & \textbf{39.5} \\
			\midrule
			\textit{Target} & \textit{39.7} & \textit{52.0} & \textit{35.4} & \textit{79.1} & \textit{74.2} & \textit{82.0} & \textit{23.1} & \textit{46.1} & \textit{60.1} & \textit{52.4} & \textit{54.4} \\
			\bottomrule
			\toprule
			\multicolumn{12}{c}{Source: \textbf{GFT}. Target: \textbf{BP4D}. } \\
			\midrule
			\textit{Source} & \textit{42.6} & \textit{35.6} & \textit{46.2} & \textit{69.8} & \textit{60.6} & \textit{77.7} & \textit{19.9} & \textit{43.3} & \textit{31.8} & \textit{46.4} & \textit{47.4}\\
			\midrule
			DANN~\cite{ganin2015unsupervised} & 12.7 & 17.3 & 10.3 & 53.5 & 50.8 & 61.7 & 5.9 & 22.4 &  32.1 & 36.0 & 30.3\\
			
			MCD~\cite{saito2018maximum} & 19.0 & 11.7 & 11.6 & 60.3 & 52.3 & 63.2 & 14.3 & 17.6 & \textbf{35.4} & 31.7 & 31.7\\
			
			P-MCD~\cite{yin2021self} & 19.2 & 11.6 & 18.5 & 61.5 & 50.3 & 68.0 & 14.0 & 19.4 & 35.1 & 39.5 & 33.7\\
			\midrule
			TCAE ~\cite{li2019self} & 26.3 & 30.7 & 20.6 & 57.6 & 52.8 & 65.6 & 10.7 & 28.2 & 24.0 & 37.1 & 35.4\\
			DGNet++~\cite{zou2020joint} & 27.3 & 29.4 & 38.4 & 54.8 & 51.1 & 53.8 & 20.1 & 34.6 & 24.2 & 33.9 & 36.8\\
			
			DRANet~\cite{lee2021dranet} & 29.0 & 32.0 & 19.8 & 63.5 & 52.0 & 69.7 & 13.3 & 33.4 & 33.4 & 41.4 & 38.8\\
			
			GCL~\cite{chen2021joint} & 30.8 & 28.8 & 40.1 & 52.5 & 46.2 & 57.3 & 16.0 & 36.6 & 32.0 & 45.6 & 38.6\\
			
			DTA~\cite{lee2019drop} & 24.2 & 19.6 & 28.7 & 52.3 & 49.5 & 70.3 & 10.4 & 22.6 & 20.8 & 26.5 & 32.5\\
			
			\textbf{AUDD(Ours)} & \textbf{47.1} & \textbf{36.5} & \textbf{49.9} & \textbf{76.6} & \textbf{63.6} & \textbf{83.8} & \textbf{27.9} & \textbf{44.1} & 32.1 & \textbf{48.9} & \textbf{51.0}\\
			\midrule
			\textit{Target} & \textit{56.1} & \textit{50.1} & \textit{51.6} & \textit{76.6} & \textit{80.8} & \textit{84.6} & \textit{55.5} & \textit{47.3} & \textit{42.7} & \textit{52.6} & \textit{59.8} \\
			\bottomrule
	\end{tabular}}
\end{table*}

\textbf{Training and evaluation protocol.} In our cross-domain experiments, we organize four datasets into four pairs (BP4D vs. GFT, BP4D+ vs. GFT, BP4D vs. DISFA, and BP4D+ vs. DISFA), where each pair alternates between serving as the source and target domains.
During the training phase, the methodology leverages labeled images from the source domain in conjunction with unlabeled images from the target domain to facilitate unsupervised cross-domain adaptation. 
The GFT dataset is utilized in accordance with its official partitioning, while the DISFA dataset is divided into three subject-independent folds to serve the purposes of training, validation, and testing, respectively.
For the purpose of evaluation within the contexts of BP4D and DISFA, the analysis concentrates on five mutually shared AUs, specifically AU1, AU2, AU4, AU6, and AU12. Conversely, the evaluation encompassing BP4D and GFT focuses on a broader set of ten shared AUs: AU1, AU2, AU4, AU6, AU10, AU12, AU14, AU15, AU23, and AU24.  
We use F1 score ($F1=\frac{2RP}{R+P}$) to measure the AU detection performance for the proposed AUDD as well as the compared methods, where $R$ and $P$ denote recall and precision, respectively. In addition, we calculate the average performance over all AUs (AVE).

\begin{table*}[ht]
	\centering
	\caption{F1-scores of domain adaptation between BP4D+ and GFT. \textbf{Bold} means the best.}
	\label{tab:4_bp4d_plus_gft}
	\scalebox{0.95} {\begin{tabular}{c|c|c|c|c|c|c|c|c|c|c|c}
			\toprule
			AU & 1 & 2 & 4 & 6 & 10 & 12 & 14 & 15 & 23 & 24 & AVE\\
			\toprule
			\multicolumn{12}{c}{Source: \textbf{BP4D+}. Target: \textbf{GFT}. } \\
			\midrule
			\textit{Source} & \textit{23.2} & \textit{28.2} & \textit{14.6} & \textit{65.4} & \textit{51.4} & \textit{69.6} & \textit{6.2} & \textit{35.4} & \textit{45.7} & \textit{37.4} & \textit{37.7}\\
			
			\midrule
			DANN~\cite{ganin2015unsupervised} & 8.3 & 12.9 & 10.7 & 54.9 & 48.7 & 56.1 & \textbf{7.0} & 8.8 & 33.8 & 16.9 & 25.9\\
			MCD~\cite{saito2018maximum} & 5.1 & 15.8 & 9.5 & 55.5 & 48.2 & 61.0 & 6.6 & 14.9 & 31.0 & 25.8 & 27.3\\
			P-MCD~\cite{yin2021self} & 7.1 & 16.9 & 10.4 & 54.0 & 54.6 & 59.1 & 6.1 & 14.5 & 40.4 & 31.8 & 29.5\\
			\midrule
			TCAE ~\cite{li2019self} & 11.5 & 27.1 & 11.2 & 65.3 & 66.3 & 66.9 & 5.4 & 25.8 & 38.8 & 32.3 & 35.1\\
			DGNet++~\cite{zou2020joint} & 11.1 & 20.9 & 5.9 & 61.2 & 68.5 & 67.6 & 2.6 & 25.4 & 41.8 & 31.9 & 33.7\\
			DRANet~\cite{lee2021dranet} & 14.8 & 27.3 & 13.7 & 61.5 & 56.8 & 66.9 & 4.2 & 14.4 & 33.2 & 30.7 & 32.4\\
			GCL~\cite{chen2021joint} & 13.7 & 29.1 & 5.9 & 62.6 & 54.7 & 64.1 & 4.2 & 24.5 & 40.3 & 28.1 & 32.7\\
			
			DTA~\cite{lee2019drop} & 8.6 & 21.8 & 6.6 & 43.4 & 39.6 & 45.1 & 5.9 & 17.4 & 20.0 & 24.8 & 23.3\\
			
			\textbf{AUDD(Ours)} & \textbf{30.4} & \textbf{33.2} & \textbf{14.8} & \textbf{70.8} & \textbf{69.9} & \textbf{78.9} & 5.4 & \textbf{35.8} & \textbf{49.9} & \textbf{38.8} & \textbf{42.8} \\
			\midrule
			\textit{Target} & \textit{39.7} & \textit{52.0} & \textit{35.4} & \textit{79.1} & \textit{74.2} & \textit{82.0} & \textit{23.1} & \textit{46.1} & \textit{60.1} & \textit{52.4} & \textit{54.4} \\
			\bottomrule
			\toprule
			\multicolumn{12}{c}{Source: \textbf{GFT}. Target: \textbf{BP4D+}. } \\
			\midrule
			\textit{Source} & \textit{28.0} & \textit{18.2} & \textit{18.6} & \textit{70.7} & \textit{73.2} & \textit{80.9} & \textit{27.2} & \textit{34.8} & \textit{42.7} & \textit{17.7} & \textit{41.2}\\
			
			\midrule
			DANN~\cite{ganin2015unsupervised} & 12.6 & 10.6 & 1.0 & 61.2 & 63.5 & 71.7 & 3.3 & 18.4 & 30.5 & 12.8 & 28.6\\
			MCD~\cite{saito2018maximum} & 16.5 & 12.2 & 10.3 & 68.2 & 68.9 & 75.8 & 8.9 & 21.5 & 32.8 & 12.6 & 32.8\\
			P-MCD~\cite{yin2021self} & 16.6 & 14.5 & 12.2 & 68.3 & 67.1 & 75.1 & 7.5 & 22.9 & 35.1 & 14.3 & 33.4\\
			\midrule
			TCAE ~\cite{li2019self} & 20.3 & 17.8 & 5.9 & 52.9 & 72.6 & 72.8 & 9.1 & 24.5 & 24.0 & 15.1 & 31.5\\
			DGNet++~\cite{zou2020joint} & 17.5 & 15.4 & 6.9 & 66.7 & 67.2 & 74.4 & 23.7 & 31.6 & 29.4 & 11.8 & 34.5\\
			DRANet~\cite{lee2021dranet} & 20.8 & 17.8 & 9.0 & 61.0 & 64.2 & 71.3 & 9.2 & 36.3 & 40.0 & 21.1 & 35.1\\
			GCL~\cite{chen2021joint} & 15.5 & 16.9 & 15.0 & 60.7 & 70.6 & 69.3 & 4.4 & 34.4 & 39.0 & \textbf{21.7} & 34.8\\
			
			DTA~\cite{lee2019drop} & 18.9 & 15.5 & 8.9 & 67.3 & 77.9 & 73.3 & 19.2 & 34.6 & 30.3 & 19.4 & 36.5\\
			
			\textbf{AUDD(Ours)} & \textbf{30.8} & \textbf{20.4} & \textbf{19.8} & \textbf{82.0} & \textbf{79.3} & \textbf{86.0} & \textbf{38.9} & \textbf{41.2} & \textbf{43.2} & 19.3 & \textbf{46.1} \\
			\midrule
			
			\textit{Target} & \textit{42.6} & \textit{35.5} & \textit{29.4} & \textit{85.4} & \textit{89.8} & \textit{87.9} & \textit{77.9} & \textit{43.0} & \textit{55.2} & \textit{28.5} & \textit{57.5} \\
			\bottomrule
	\end{tabular}}
	
\end{table*}

\begin{table}[ht]
	\centering
	\caption{F1-scores of domain adaptation between BP4D and DISFA. \textbf{Bold} means the best.}
	\label{tab:4_bp4d_disfa}
	\scalebox{0.9} {\begin{tabular}{c|c|c|c|c|c|c}
			\toprule
			AU & 1 & 2 & 4 & 6 & 12 & AVE\\
			\toprule
			\multicolumn{7}{c}{Source: \textbf{BP4D}. Target: \textbf{DISFA}. } \\
			\midrule
			\textit{Source} & \textit{27.7} & \textit{28.0} & \textit{52.2} & \textit{49.5} & \textit{62.0} & \textit{43.9} \\
			\midrule
			DANN~\cite{ganin2015unsupervised} & 10.1 & 8.1 & 17.1 & 33.3 & 56.8 & 25.1\\
			MCD~\cite{saito2018maximum} & 17.9 & 18.1 & 31.5 & 30.2 & 38.9 & 27.3\\
			P-MCD~\cite{yin2021self} & 34.3 & 16.6 & 52.1 & 33.5 & 50.4 & 37.4\\
			\midrule
			TCAE~\cite{li2019self} & 22.2 & 37.6 & 42.1 & 32.7 & 61.9 & 39.3\\
			DGNet++~\cite{zou2020joint} & 17.9 & 9.6 & 38.0 & 28.2 & 64.8 & 31.7\\
			DRANet~\cite{lee2021dranet} & 33.1 & 27.0 & 49.2 & 33.4 & 65.7 & 41.7\\
			GCL~\cite{chen2021joint} & 25.8 & 34.6 & 46.0 & 33.6 & \textbf{69.7} & 41.9\\
			
			DTA~\cite{lee2019drop} & 14.4 & 11.7 & 21.7 & 20.6 & 37.5 & 21.2\\
			
			\textbf{AUDD(Ours)} & \textbf{37.8} & \textbf{38.1} & \textbf{56.3} & \textbf{53.0} & 67.6 & \textbf{50.6} \\
			\midrule
			\textit{Target} & \textit{44.6} & \textit{41.4} & \textit{64.2} & \textit{59.8} & \textit{75.2} & \textit{57.0}\\
			\bottomrule
			\toprule
			\multicolumn{7}{c}{Source: \textbf{DISFA}. Target: \textbf{BP4D}. } \\
			\midrule
			\textit{Source} & \textit{54.1} & \textit{40.7} & \textit{46.1} & \textit{69.7} & \textit{74.3} & \textit{57.0} \\
			\midrule
			DANN~\cite{ganin2015unsupervised} & 27.3 & 11.5 & 39.6 & 49.5 & 55.7 & 36.7\\
			MCD~\cite{saito2018maximum} & 31.1 & 24.0 & 32.6 & 66.5 & 54.5 & 41.7\\
			P-MCD~\cite{yin2021self} & 27.2 & 26.2 & 32.0 & 68.3 & 58.0 & 42.3\\
			\midrule
			TCAE~\cite{li2019self} & 39.2 & 44.7 & 32.8 & 68.5 & 57.5 & 48.5\\
			DGNet++~\cite{zou2020joint} & 33.6 & 35.7 & 41.0 & 68.6 & 65.2 & 48.8\\
			DRANet~\cite{lee2021dranet} & 45.1 & 49.7 & \textbf{52.6} & 60.9 & 51.1 & 51.9\\
			GCL~\cite{chen2021joint} & 49.9 & 48.8 & 44.1 & 62.7 & 46.5 & 50.4\\
			
			DTA~\cite{lee2019drop} & 34.1 & 29.7 & 30.0 & 52.3 & 60.3 & 41.3\\
			
			\textbf{AUDD(Ours)} & \textbf{56.9} & \textbf{52.4} & \textbf{52.6} & \textbf{74.6} & \textbf{77.4} & \textbf{62.8} \\
			\midrule
			
			\textit{Target} & \textit{58.3} & \textit{51.1} & \textit{57.0} & \textit{80.8} & \textit{84.7} & \textit{66.4}\\
			\bottomrule
	\end{tabular}}
\end{table}

\textbf{Implementation Details}.
During training, we set the actual weight for $\alpha$ and $\beta$ as $1:1$ according to the cross-domain AU detection performance on the validation set. We implemented all the experiments using PyTorch on two RTX 3090 GPUs, each with 24 GB memory. We set the respective training batch size $N=36$ for the source and target images, respectively. We use the commonly-used data augmentation strategies, including randomly data cropping, random horizontal flipping. The neural architecture, i.e., ResNet50-Vit-B/16~\cite{dosovitskiy2020image} was pretrained on ImageNet-21K. We set the learning rate as $0.001$ and trained the AUDD model for 50 epochs until convergence for each cross-domain setting. We use the popular \textit{thop}\footnote{ PyTorch-OpCounter: \url{https://github.com/Lyken17/pytorch-OpCounter}}  library to measure the number of parameters and FLOPs of the utilized AU encoder. The AU encoder comprises 97.90 million parameters and requires 16.83 GFLOPs.

\subsection{Comparison with State-of-the-arts}
\label{sec:exp_sota}
We compare AUDD with the representative unsupervised domain adaptation methods for cross-domain AU detection,  including DANN~\cite{ganin2015unsupervised}, MCD~\cite{saito2018maximum}, P-MCD~\cite{yin2021self}.
We also compare AUDD with three representative feature-decoupling-based UDA methods, including DG-Net++~\cite{zou2020joint}, DRANet~\cite{lee2021dranet}, GCL~\cite{chen2021joint}. Notably, DG-Net++ and GCL were originally developed for cross-domain person re-identification but were adapted for cross-domain AU detection based on their feature disentangling principle. They can be naturally used for cross-domain AU detection following their feature disentangling principle. For a fair comparison with AUDD,  we re-trained these three models using the FACS datasets with the released codes. Besides, we compare AUDD with the another dropout-based DA method: Drop to Adapt (DTA)~\cite{lee2019drop}, which leverages adversarial dropout to learn discriminative features by enforcing the cluster assumption.
We also compare AUDD with the self-supervised AU detection method TCAE~\cite{li2019self}, where a pre-trained AU detection model is publicly available. Following TCAE~\cite{li2019self}, we train a linear classifier with the pre-trained model and evaluate its generalization capability on the target FACS dataset.

We show the quantitative cross-domain AU detection comparison in Tab.~\ref{tab:4_bp4d_gft}, Tab.~\ref{tab:4_bp4d_plus_gft}, Tab.~\ref{tab:4_bp4d_disfa} and Tab.~\ref{tab:4_bp4d_plus_disfa}. In all the tables, we use the ``Source'' and ``Target'' as the indicators for the lower and upper bounds. ``Source'' means directly evaluating the source model on the target domain. ``Target'' means within-domain training and evaluation on the target dataset. 

From the quantitative results in Tab.~\ref{tab:4_bp4d_gft}, Tab.~\ref{tab:4_bp4d_plus_gft}, Tab.~\ref{tab:4_bp4d_disfa} and Tab.~\ref{tab:4_bp4d_plus_disfa}, we have three observations: (1) Compared with the general domain adaptation methods including DANN~\cite{ganin2015unsupervised}, MCD~\cite{saito2018maximum} and P-MCD~\cite{yin2021self}, our proposed AUDD obtains consistent improvements in the average F1 scores.  The observed improvements suggest that the AUDD-learned AU features are more cross-domain generalizable and also indicate the effectiveness the proposed channel-wise and token-wise feature dropping mechanism. 
(2) Compared with the feature-disentangling-based methods, e.g., TCAE~\cite{li2019self}, DGNet++~\cite{zou2020joint}, DRANet~\cite{lee2021dranet} and GCL~\cite{chen2021joint}, that builds their cross-domain AU detection performance on the quality of the generated faces, our proposed AUDD is more straightforward and learns the generalizable features in a discriminative manner via incorporating the prior AU relation knowledge into the framework.
(3) Compared with DTA~\cite{lee2019drop}, AUDD consistently illustrates its superiority under various cross-domain settings. As a comparison, AUDD is assumption-free and can be trained straightforwardly without alternating adversarial manner. 

Overall, the improvements of AUDD over other compared methods should be reasonable because AUDD is capable of dropping and removing the domain-sensitive features both in the shallow and the deep layers.  Besides, it specially considers the intrinsic AUs relation during training. We will explore the benefits of each of the components in AUDD in Sec.~\ref{sec:exp_ablation_study}.

\begin{figure}[ht]
	\centering
	\begin{subfigure}
		\centering
		\includegraphics[width=1.0\linewidth]{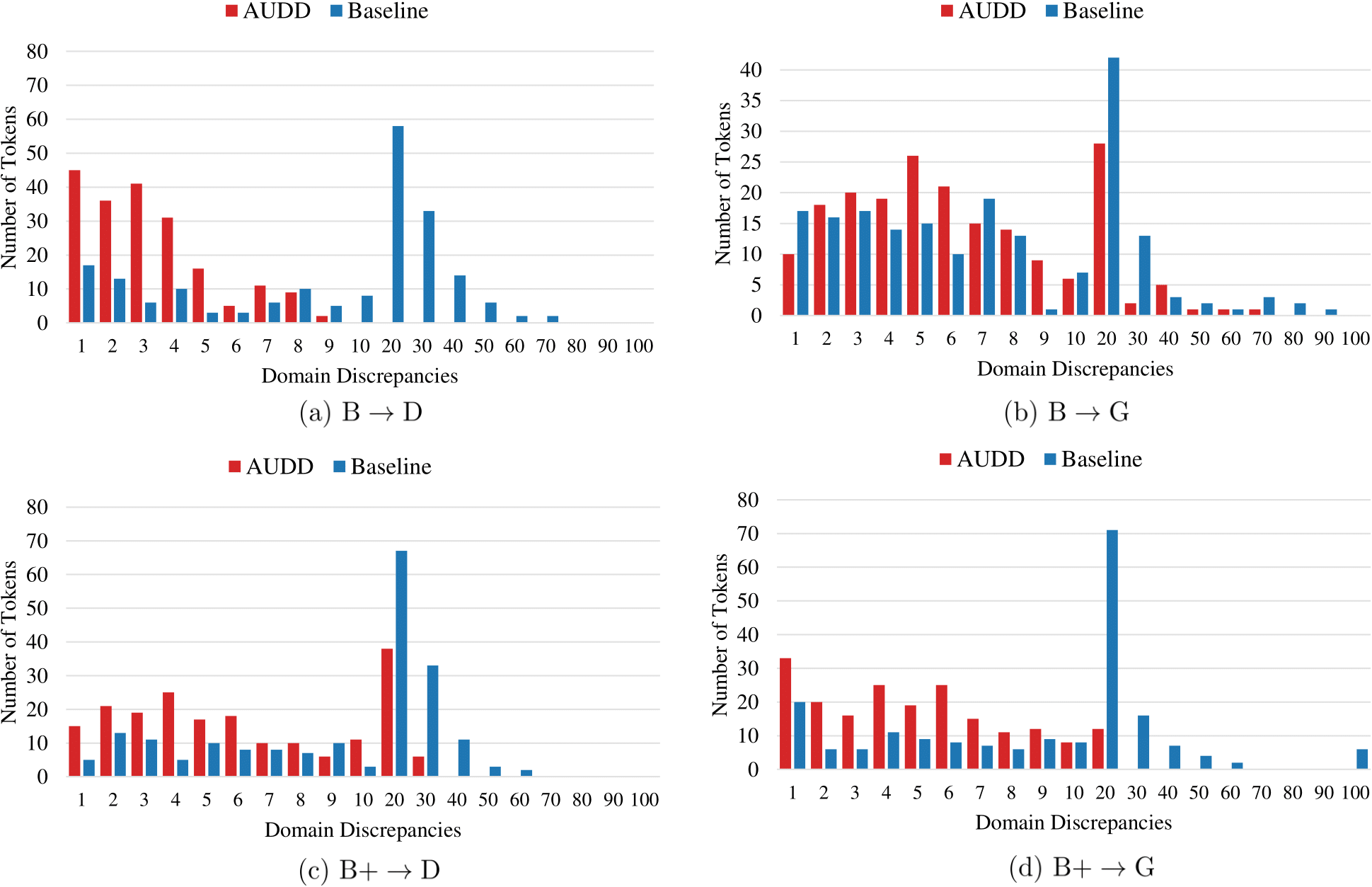}
		\label{fig:1}
	\end{subfigure}%
	
	\begin{subfigure}
		\centering
		\includegraphics[width=1.0\linewidth]{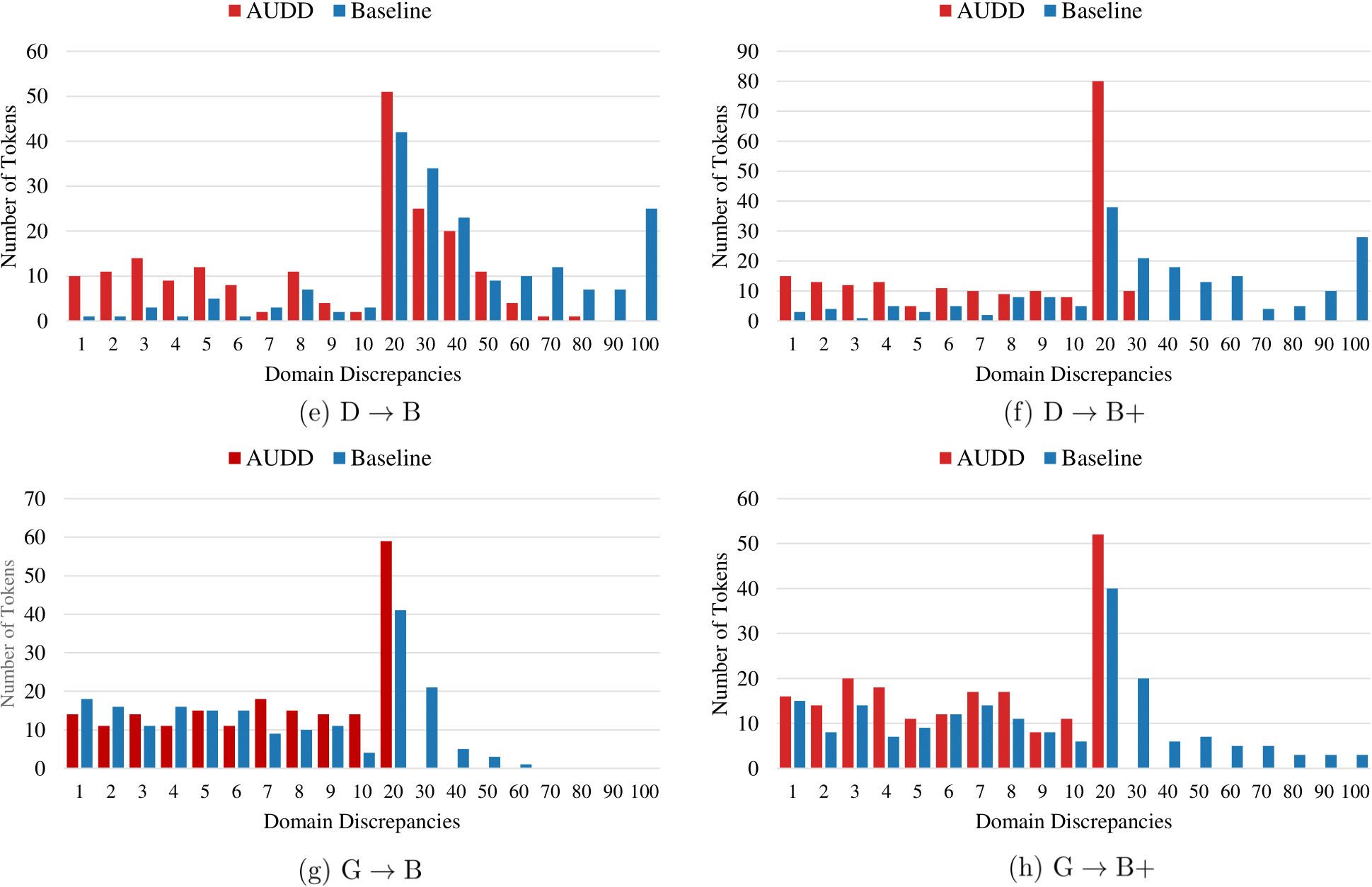}
		\label{fig:2}
	\end{subfigure}%
	\hspace{.2in}
	\caption{Domain discrepancy comparisons between AUDD and Baseline method under various cross-domain settings. It is obvious  reveal that features encoded by AUDD consistently exhibit lower sensitivity to domain shifts.}
	\label{fig:b-g}
\end{figure}

\begin{table}[ht]
	\centering
	\caption{F1-scores of domain adaptation between BP4D+ and DISFA. \textbf{Bold} means the best.}
	\label{tab:4_bp4d_plus_disfa}
	\scalebox{0.95} {\begin{tabular}{c|c|c|c|c|c|c}
			\toprule
			AU & 1 & 2 & 4 & 6 & 12 & AVE\\
			\toprule
			\multicolumn{7}{c}{Source: \textbf{BP4D+}. Target: \textbf{DISFA}. } \\
			\midrule
			\textit{Source} & \textit{29.5} & \textit{23.2} & \textit{49.9} & \textit{42.0} & \textit{68.7} & \textit{42.7} \\
			\midrule
			DANN~\cite{ganin2015unsupervised} & 33.4 & 22.7 & 43.2 & 24.0 & 35.8 & 31.8\\
			MCD~\cite{saito2018maximum} & 18.4 & 18.7 & 37.7 & 28.4 & 57.5 & 32.1\\
			P-MCD~\cite{yin2021self} & 25.2 & 10.9 & 56.5 & 37.0 & 59.5 & 37.8\\
			\midrule
			TCAE~\cite{li2019self} & 30.7 & 21.5 & 38.5 & 31.4 & 55.2 & 35.5\\
			DGNet++~\cite{zou2020joint} & 39.0 & \textbf{28.2} & 25.3 & 39.2 & 59.1 & 38.2\\
			DRANet~\cite{lee2021dranet} & \textbf{44.8} & 24.5 & 47.8 & 41.7 & 66.7 & 45.1\\
			GCL~\cite{chen2021joint} & 43.5 & 17.2 & 53.6 & 37.6 & 65.2 & 43.4\\
			
			DTA~\cite{lee2019drop} & 14.4 & 11.7 & 21.7 & 20.5 & 37.5 & 21.2\\
			
			\textbf{AUDD(Ours)} & 39.1 & 25.4 & \textbf{62.6} & \textbf{47.9} & \textbf{70.6} & \textbf{49.1} \\
			\midrule
			\textit{Target} & \textit{44.6} & \textit{41.4} & \textit{64.2} & \textit{59.8} & \textit{75.2} & \textit{57.0}\\
			\bottomrule
			\toprule
			\multicolumn{7}{c}{Source: \textbf{DISFA}. Target: \textbf{BP4D+}. } \\
			\midrule
			\textit{Source} & \textit{29.8} & \textit{24.2} & \textit{14.1} & \textit{79.2} & \textit{78.9} & \textit{45.2} \\
			\midrule
			DANN~\cite{ganin2015unsupervised} & 14.7 & 9.8 & 11.1 & 61.1 & 67.9 & 32.9\\
			MCD~\cite{saito2018maximum} & 15.2 & 14.4 & 9.7 & 60.6 & 68.6 & 33.7\\
			P-MCD~\cite{yin2021self} & 15.6 & 14.7 & 17.7 & 60.7 & 74.5 & 36.6\\
			\midrule
			TCAE~\cite{li2019self} & 21.8 & 18.4 & 16.0 & 52.1 & 72.8 & 36.2\\
			DGNet++~\cite{zou2020joint} & 12.1 & 15.5 & 14.2 & 70.6 & 74.3 & 37.3\\
			DRANet~\cite{lee2021dranet} & 23.3 & 15.8 & \textbf{20.2} & 70.8 & 68.9 & 39.8\\
			GCL~\cite{chen2021joint} & 23.9 & 18.1 & 16.2 & 76.1 & 69.1 & 40.7\\
			
			DTA~\cite{lee2019drop} & 18.9 & 15.0 & 8.9 & 32.8 & 46.3 & 24.4\\
			
			\textbf{AUDD(Ours)} & \textbf{35.7} & \textbf{27.2} & 18.4 & \textbf{80.6} & \textbf{86.4} & \textbf{49.7} \\
			\midrule
			
			\textit{Target} & \textit{41.8} & \textit{38.8} & \textit{31.3} & \textit{85.0} & \textit{88.4} & \textit{57.1}\\
			\bottomrule
	\end{tabular}}
\end{table}

To evaluate the robustness of the proposed AUDD model against domain shifts between the source and target domains, the output from the terminal layer of the backbone network serves as representations of action units (AUs). Subsequently, we assess the feature disparities between the source and target domains.
As there are $N = 768$ tokens,  we calculate the average activation values for $i-$th token w.r.t the source/target domain as: $a_{i}^{k}=\frac{1}{n_k} \sum_{j=1}^{n_k} \text{GAP}(\mathbf{x}^{j,i}_3)$, where $\mathbf{x}^{j,i}_3$ denotes the $i$-th token generated by the third transformer block w.r.t the $j$-th input image. $n_{k}$ means the total images in the source ($n_{s}=N$) or target ($n_{t}=M$) domain. Then we can obtain the domain discrepancies between the source and target domains as: $D_{i} = \frac{| a_{i}^{s} - a_{i}^{t} |}{\max(| a_{i}^{s} |, | a_{i}^{t}) |} \times 100$. Here we exploit the maximal values for numeric normalization. Finally, we visualize the cross-domain discrepancies under different cross-domain settings in Fig.~\ref{fig:b-g}.
The comparisons presented in Fig.~\ref{fig:b-g} demonstrate that our proposed AUDD consistently exhibits fewer tokens with noticeable domain discrepancies compared to the baseline method. Moreover, most tokens in AUDD demonstrate lower discrepancies than those in the baseline method.
The comparisonS depicted in Fig.~\ref{fig:b-g} suggest that AUDD effectively identifies and removes domain-sensitive tokens during the training process. Tokens in AUDD contain less domain-dependent information, highlighting the consistent efficacy of the proposed selective dropout method.

\begin{figure}[ht]
	\centering
	\includegraphics[width=1.0\linewidth]{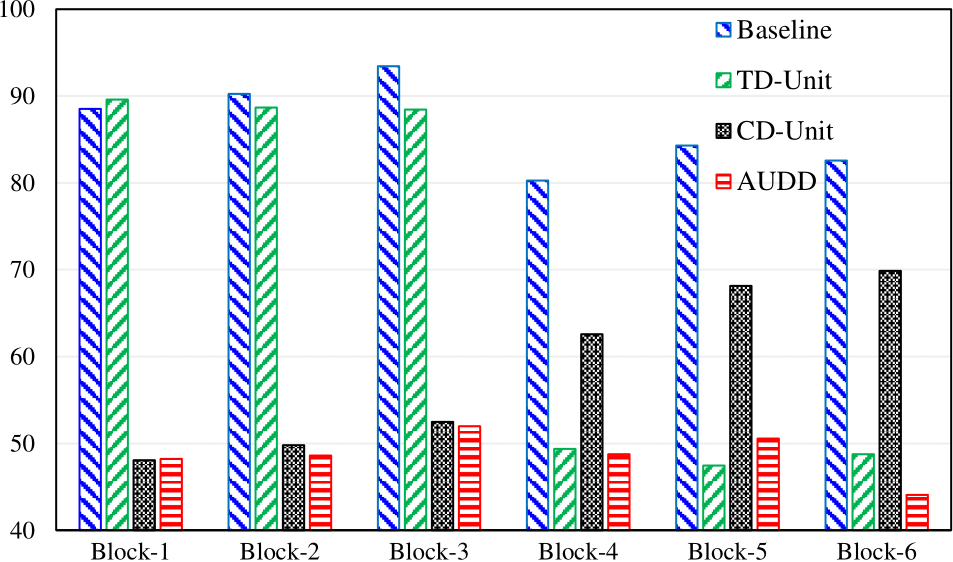}
	\caption{Domain classification performance w.r.t. different blocks (Source: BP4D, Target:DISFA). AUDD shows consistent low domain classification accuracy, indicating the domain-related features have been well mitigated.}
	\label{fig:domain-acc}
\end{figure}

\subsection{Analysis of Each Component in AUDD}
\label{sec:exp_ablation_study}
\textbf{Quantitative analysis.} Firstly, we verify the binary domain classification performance w.r.t. different blocks in the backbone under different methods. The comparison is shown in Fig.~\ref{fig:domain-acc}. Firstly, the vanilla baseline method show quite high domain classification accuracy under various blocks, suggesting the domain-related features reside in both the shallow and the deep layers within the backbone network. Secondly, CD-Unit helps the model to mitigate the domain-sensitive features in the preceding CNN parts, while TD-Unit consistently and merely removes the domain-dependent features in the following transformer parts. Thirdly, our proposed AUDD shows consistent low domain classification accuracy for different layers, indicating the CD-Unit and TD-Unit mechanisms work in a collaborative manner.

Secondly, we perform extra experiments to investigate the contribution of each component in our proposed AUDD, i.e., CD-Unit, TD-Unit, as well as the vanilla domain adversarial learning method. We show the numeric results and comparisons in Tab.~\ref{tab:4_ablation_study}. 

\begin{figure*}[htb]
	\centering
	\includegraphics[width=0.8\linewidth]{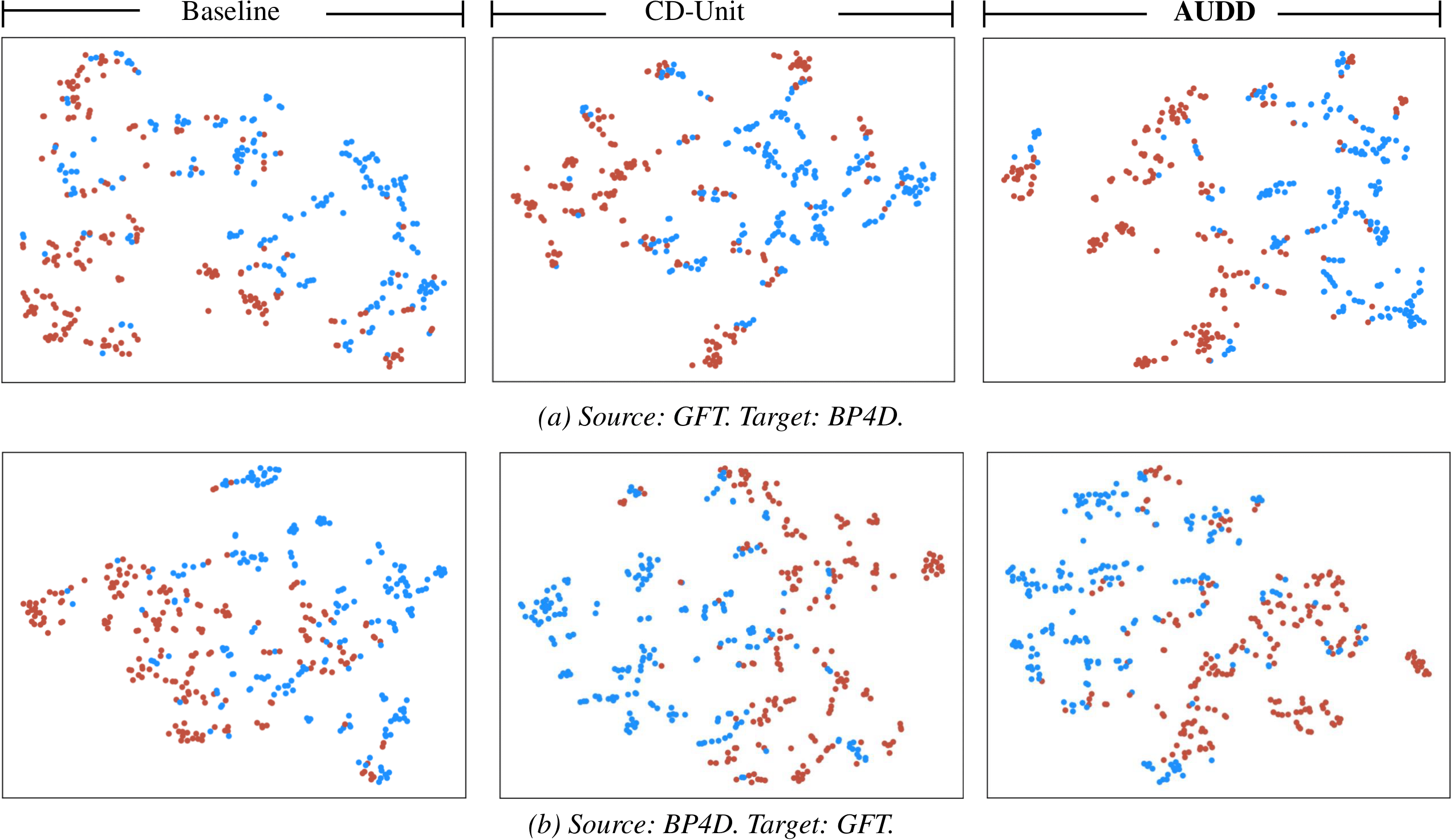}
	\caption{t-SNE visualization of the learned features. AUDD shows the best feature discrimination w.r.t AU6. \textcolor{redcolor}{Red}/\textcolor{bluecolor}{Blue} indicate AU6 exists or not.}
	\label{fig:4-tsne_b2g}
\end{figure*}

\begin{table}[ht]
	\centering
	\caption{Cross-dataset AU detection performance under different configurations. \textbf{Bold} denotes best. \underline{Underline} denotes second best.
	}
	\label{tab:4_ablation_study}
	\scalebox{0.95}{
		\begin{tabular}{c|c|c|c|c|c}
			\toprule
			Datasets & Baseline & Baseline$+$DA & CD-Unit & TD-Unit & \textbf{AUDD} \\
			\midrule
			B $\rightarrow$ G & 36.9 & 37.3 & 38.2 & \underline{38.3} & \textbf{39.5} \\
			G $\rightarrow$ B & 47.4 & 49.2 & 50.0 & \underline{50.3} & \textbf{51.0} \\
			B+ $\rightarrow$ G & 37.7 & 39.9 & 40.6 & \underline{41.0} & \textbf{42.8} \\
			G $\rightarrow$ B+ & 41.2 & 43.3 & 44.0 & \underline{44.6} & \textbf{46.1} \\
			B $\rightarrow$ D & 43.9 & 45.1 & 47.3 & \underline{48.2} & \textbf{50.6} \\
			D $\rightarrow$ B & 57.0 & 57.8 & 59.5 & \underline{60.1} & \textbf{62.8} \\
			B+ $\rightarrow$ D & 42.7 & 44.8 & 46.2 & \underline{46.7} & \textbf{49.1} \\
			D $\rightarrow$ B+ & 45.2 & 47.5 & \underline{48.6} & 48.1 & \textbf{49.7} \\
			\bottomrule
		\end{tabular}
	}
\end{table}

\begin{figure}[ht]
	\centering
	\centering
	\includegraphics[width=1.0\linewidth]{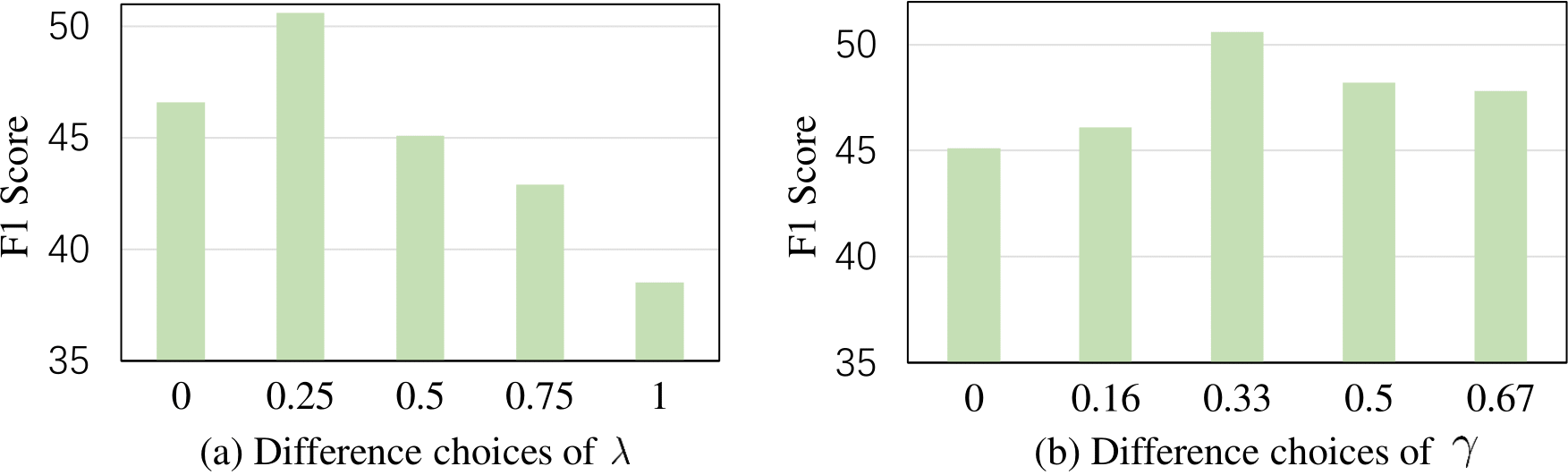}
	\caption{Performance variations under different choice of $\lambda$ and $\gamma$ (Source: BP4D, Target:DISFA).}
	\label{fig:4_GRLRate}
\end{figure}

\begin{figure*}[ht]
	\centering
	\centering
	\includegraphics[width=0.8\linewidth]{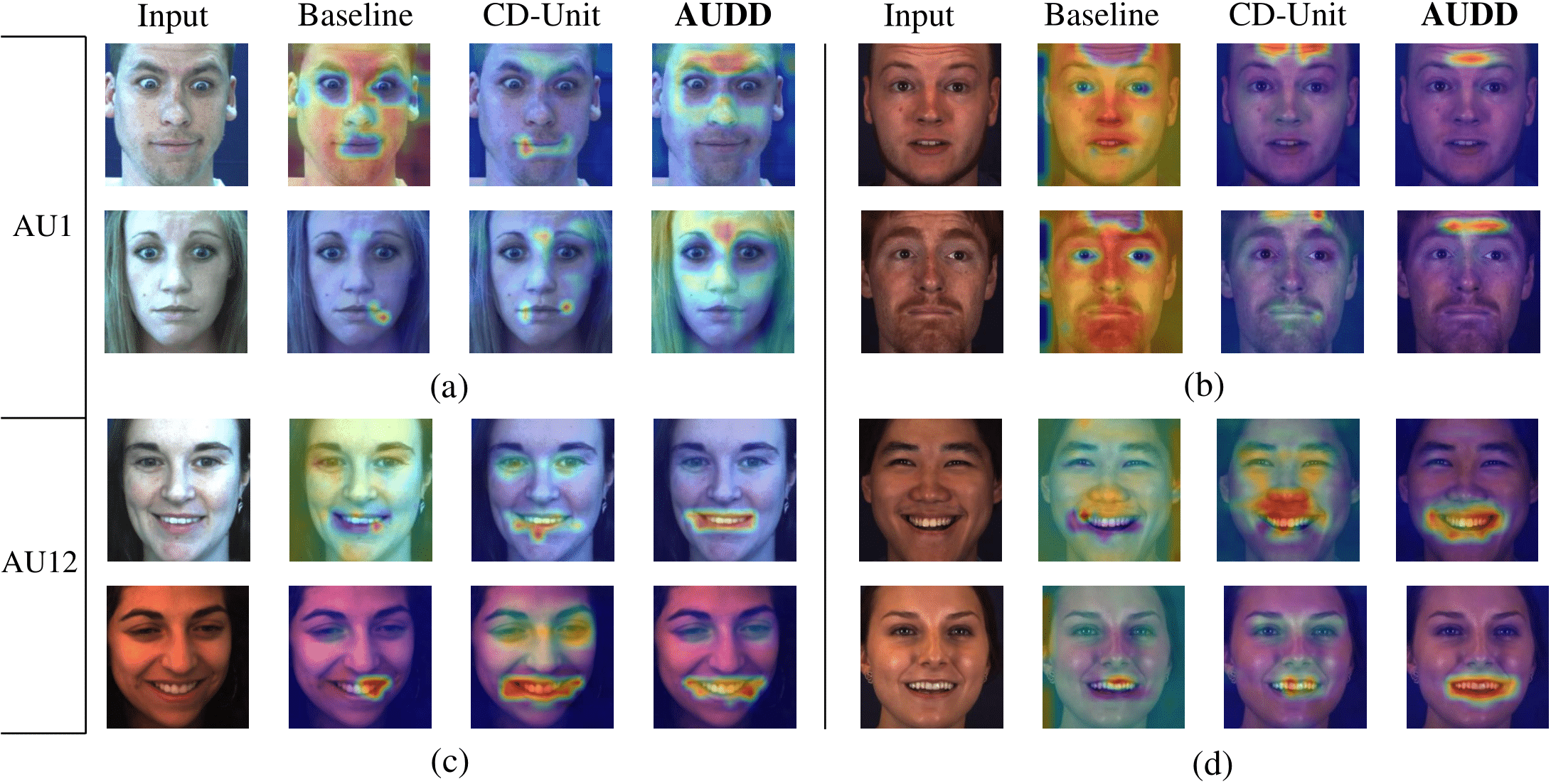}
	\caption{Visualization of attention maps of the baseline and the compared methods. In sub-figures (a) and (c), BP4D is designated as the source dataset and DISFA as the target. Conversely, in sub-figures (b) and (d), the roles of the datasets are reversed. It is distinctly apparent that the attention maps produced by AUDD exhibit a more concentrated focus concerning the relevant AU.}
	\label{fig:individual_au}
\end{figure*}

\begin{figure*}[htb]
	\centering
	\centering
	\includegraphics[width=0.8\linewidth]{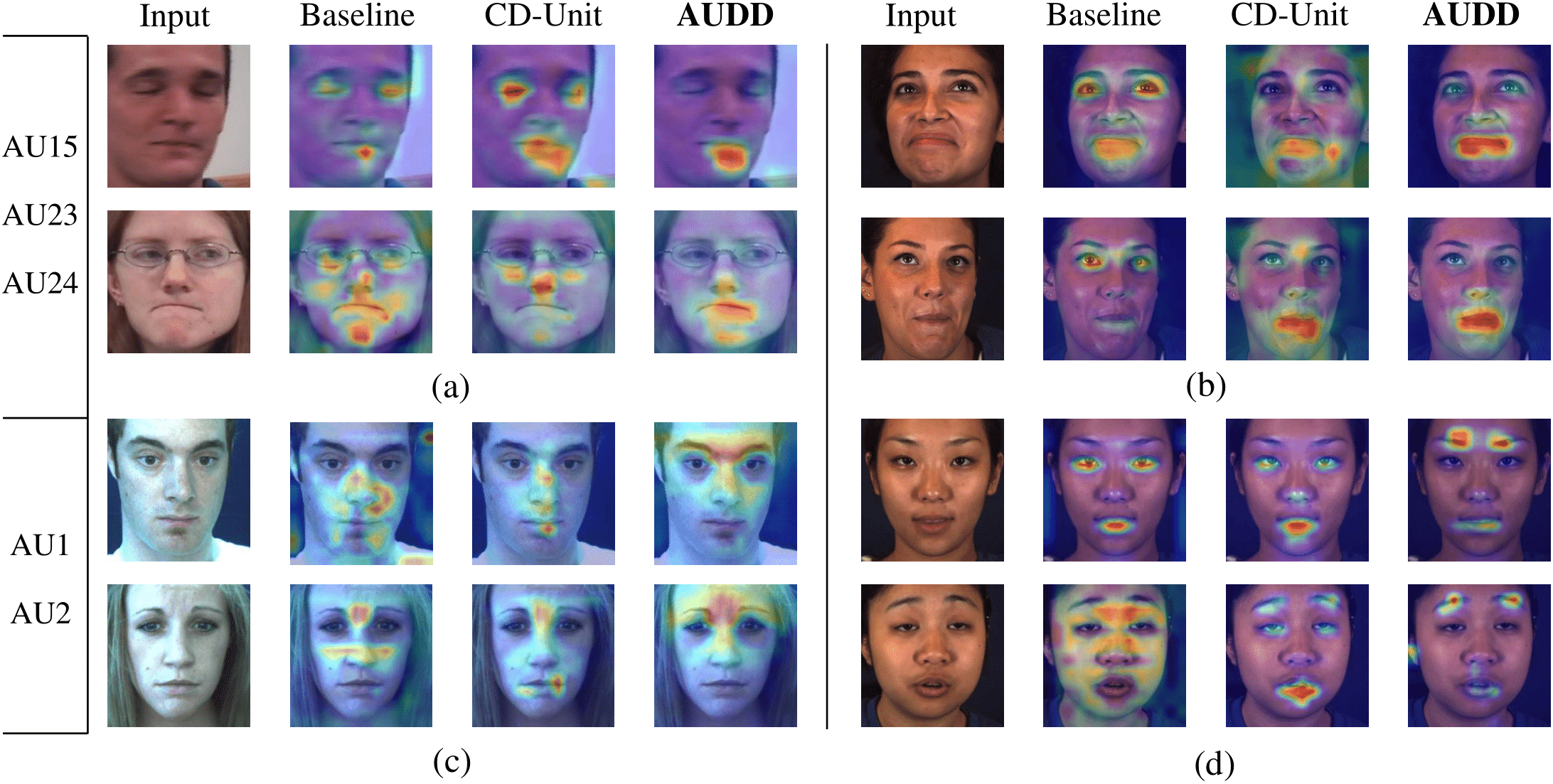}
	\caption{Visualization of attention maps of the baseline and the compared methods. The sub-figure (a) delineates the attention maps with BP4D+ as the source dataset and GFT as the target, whereas (b) illustrates the inverse configuration. (c) denotes the source and target dataset as BP4D and DISFA. For (d), the vice versa. It is evident that the attention maps produced by AUDD show enhanced focus with respect to the pertinent AUs.}
	\label{fig:combinational_au}
\end{figure*}

In Tab.~\ref{tab:4_ablation_study}, the Baseline method means we merely use the vanilla ResNet50-ViT-B/16 backbone. ``Baseline+DA'' means we use multi-layer domain adversarial learning with the backbone. ``CD-Unit'' means we merely exploit progressive channel-wise selective dropout.
``TD-Unit'' means we merely exploit progressive token-wise selective dropout. We have four observations: (1) The introduction of multi-layer domain adversarial learning leads to minor enhancements (as seen when comparing column 2 with column 3). This is understandable, given that the domain discriminator's role is to align the features from source and target domains without taking class labels into account. Consequently, while the features may become domain-invariant, the preservation of class-discriminative properties is not assured.
(2) Both CD-Unit and TD-Unit are effective. They both show more obvious improvements than merely adding domain adversarial learning, indicating the importance of domain-sensitive feature dropping. The efficacy of the CD-Unit lies in its ability to eliminate domain-sensitive channel-wise features, each representing a unique pattern, while the TD-Unit aids the model in discerning the inherent relationships among AUs, thus enhancing cross-domain adaptability.
(3) The TD-Unit (column 5) exhibits marginal superiority over the CD-Unit (column 4). This finding aligns with insights from ~\cite{yosinski2014transferable}, suggesting that although both shallow and deep network layers harbor domain-relevant information, the deeper layers are more likely to be task-specific and may harbor a greater amount of domain-specific content.
(4) AUDD obtains the best results under various different cross-domain AU detection settings, indicating CD-Unit and TD-Unit collaboratively drop the domain-sensitive features from different aspects.

\textbf{Feature visualization.} We show the feature distributions using our proposed AUDD and the vanilla Baseline method in Fig.~\ref{fig:4-tsne_b2g}. 
For each sub-figure in Fig.~\ref{fig:4-tsne_b2g}, totally 400 facial images were randomly sampled from the testing set of the target dataset, with half of the images show activated AU6. The features of the selected facial images are projected into a 2D space by t-SNE.
It is obvious that incorporating the CD-Unit will achieve more discriminative AU features in the target dataset, indicating more cross-dataset generalizability. Besides, our proposed AUDD (the rightmost columns) shows the best  feature discrimination w.r.t AU6. The comparison in Fig.~\ref{fig:4-tsne_b2g} further verifies the effectiveness of the proposed selective dropout learning paradigm for cross domain facial action unit detection.

\textbf{Sensitivity of hyperparameters.} Our investigation focuses on assessing the influence of two crucial hyperparameters: $\lambda$ and $\gamma$. Here, $\lambda$ represents a coefficient regulating the impact of the Gradient Reversal Layer (GRL) within the domain classifier, while $\gamma$ specifies the proportion of channels or tokens to be selectively discarded in each training iteration. The experimental findings are illustrated in Fig.~\ref{fig:4_GRLRate}.

For $\lambda$, the empirical data in Fig.~\ref{fig:4_GRLRate} (a) clearly indicates that the AUDD model attains optimal performance at $\lambda = 0.25$. This result shows the efficacy of gradient reversal operation in facilitating the model's acquisition of domain-invariant features. However, a performance decline is observed with increasing $\lambda$ values, suggesting that an excessive emphasis on learning domain-invariant features, due to a higher GRL weight, may detract from the model's ability to discern AU-specific features. Regarding $\gamma$, optimal results are achieved with a value of $\gamma = 0.33$, as shown in Fig.~\ref{fig:4_GRLRate} (b). This finding implies that a lower $\gamma$ value may be insufficient to prompt the model to effectively eliminate domain-sensitive features. Conversely, an elevated $\gamma$ value could be detrimental to performance, as the removal of an excessive number of features might pose substantial challenges to the model's training process.

\textbf{Attention map investigation.} 
We delve into the analysis of GradCAM-based attention maps for our AUDD model to extend our exploration~\cite{selvaraju2017grad}. Specifically, we present the attention maps in relation to individual AU as well as combinations thereof in Fig.~\ref{fig:individual_au} and ~\ref{fig:combinational_au}, respectively. For ease of illustration, combinational AUs are categorized into distinct groups.

We have three observations. 
Initially, the efficacy of the CD-Unit is evident, as it facilitates the model's focus on pertinent facial regions to a certain extent, as illustrated by the comparative analysis between the ``Baseline'' and ``CD-Unit'' in each sub-figure. However, the resulting attention maps from the exclusive use of the CD-Unit fail to meet expectations, suggesting that mere channel dropout in convolutional layers is inadequate. 
Secondly, the integration of the TD-Unit within the AUDD framework notably enhances the clarity of the attention maps, particularly demonstrated in Fig.~\ref{fig:individual_au}. This observation suggests a synergistic operation between the CD-Unit and TD-Unit, further augmenting the model's capability.
Thirdly, as depicted in the comparison in Fig.~\ref{fig:combinational_au}, AUDD significantly improves the attention maps concerning combinational AUs. This finding indicates that integrating the TD-Unit effectively captures AU relationships and enhances their representation, contributing to the overall efficacy of the model.

\section{Conclusion}
\label{sec:conclusion}
Within this paper we have introduced a Doubly adaptive Dropout method tailored specifically for cross-domain AU detection. Our approach integrates a Channel Drop Unit (CD-Unit) and a Token Drop Unit (TD-Unit) to effectively mitigate domain-specific disparities at both the channel and token levels. The CD-Unit is tasked to preserve domain-agnostic local patterns within the convolutional feature maps. Concurrently, the TD-Unit is devised to enable the model to discern the inherent AU relationships that exhibit a high degree of generalizability across different domains. These two units collaborate to eliminate domain-sensitive features at varying levels of granularity, aiming to enhance the model's adaptability and performance in various cross-domain settings.
One limitation of AUDD is that it does not incorporate foundational vision models or language-based AU definition to further improve the generalization of cross-domain AU detection. We will explore these in future work.
	
\section{Acknowledgment}
This research was partially supported by the RIE2020 AME Programmatic Fund, Singapore (No. A20G8b0102), the A*STAR Prenatal / Early Childhood Grant (No. H22P0M0002), the National Natural Science Foundation of China (No. 62272231), the Fundamental Research Funds for the Central Universities (No. 4009002401), and the Big Data Computing Center of Southeast University.
	
	\bibliographystyle{IEEEtran}
	\bibliography{egbib}

\begin{thebibliography}{10}
\providecommand{\url}[1]{#1}
\csname url@samestyle\endcsname
\providecommand{\newblock}{\relax}
\providecommand{\bibinfo}[2]{#2}
\providecommand{\BIBentrySTDinterwordspacing}{\spaceskip=0pt\relax}
\providecommand{\BIBentryALTinterwordstretchfactor}{4}
\providecommand{\BIBentryALTinterwordspacing}{\spaceskip=\fontdimen2\font plus
\BIBentryALTinterwordstretchfactor\fontdimen3\font minus
  \fontdimen4\font\relax}
\providecommand{\BIBforeignlanguage}[2]{{%
\expandafter\ifx\csname l@#1\endcsname\relax
\typeout{** WARNING: IEEEtran.bst: No hyphenation pattern has been}%
\typeout{** loaded for the language `#1'. Using the pattern for}%
\typeout{** the default language instead.}%
\else
\language=\csname l@#1\endcsname
\fi
#2}}
\providecommand{\BIBdecl}{\relax}
\BIBdecl

\bibitem{picard2000affective}
R.~W. Picard, \emph{Affective computing}.\hskip 1em plus 0.5em minus
  0.4em\relax MIT press, 2000.

\bibitem{friesen1978facial}
E.~Friesen and P.~Ekman, ``Facial action coding system: a technique for the
  measurement of facial movement,'' \emph{Palo Alto}, vol.~3, 1978.

\bibitem{li2018occlusion}
Y.~Li, J.~Zeng, S.~Shan, and X.~Chen, ``Occlusion aware facial expression
  recognition using cnn with attention mechanism,'' \emph{IEEE Transactions on
  Image Processing}, vol.~28, no.~5, pp. 2439--2450, 2018.

\bibitem{lucey2009automatically}
P.~Lucey, J.~Cohn, S.~Lucey, I.~Matthews, S.~Sridharan, and K.~M. Prkachin,
  ``Automatically detecting pain using facial actions,'' in \emph{2009 3rd
  International Conference on Affective Computing and Intelligent Interaction
  and Workshops}.\hskip 1em plus 0.5em minus 0.4em\relax IEEE, 2009, pp. 1--8.

\bibitem{niu2023cnn}
K.~Niu, G.~Lu, X.~Peng, Y.~Zhou, J.~Zeng, and K.~Zhang, ``Cnn autoencoders and
  lstm-based reduced order model for student dropout prediction,'' \emph{Neural
  Computing andApplications}, vol.~35, no.~30, pp. 22\,341--22\,357, 2023.

\bibitem{wang2017expression}
S.~Wang, Q.~Gan, and Q.~Ji, ``Expression-assisted facial action unit
  recognition under incomplete au annotation,'' \emph{Pattern Recognition},
  vol.~61, pp. 78--91, 2017.

\bibitem{li2021meta}
Y.~Li and S.~Shan, ``Meta auxiliary learning for facial action unit
  detection,'' \emph{IEEE Transactions on Affective Computing}, vol.~14, no.~3,
  pp. 2526--2538, 2021.

\bibitem{ghosh2015multi}
S.~Ghosh, E.~Laksana, S.~Scherer, and L.-P. Morency, ``A multi-label
  convolutional neural network approach to cross-domain action unit
  detection,'' in \emph{2015 International Conference on Affective Computing
  and Intelligent Interaction (ACII)}.\hskip 1em plus 0.5em minus 0.4em\relax
  IEEE, 2015, pp. 609--615.

\bibitem{onal2019cross}
I.~Onal~Ertugrul, J.~Cohn, L.~Jeni, Z.~Zhang, L.~Yin, and Q.~Ji, ``Cross-domain
  au detection: Domains, learning approaches, and measures,'' \emph{FG. IEEE},
  2019.

\bibitem{ertugrul2020crossing}
I.~O. Ertugrul, J.~F. Cohn, L.~A. Jeni, Z.~Zhang, L.~Yin, and Q.~Ji, ``Crossing
  domains for au coding: Perspectives, approaches, and measures,'' \emph{IEEE
  transactions on biometrics, behavior, and identity science}, vol.~2, no.~2,
  pp. 158--171, 2020.

\bibitem{yin2021self}
Y.~Yin, L.~Lu, Y.~Wu, and M.~Soleymani, ``Self-supervised patch localization
  for cross-domain facial action unit detection,'' in \emph{2021 16th IEEE
  International Conference on Automatic Face and Gesture Recognition (FG
  2021)}.\hskip 1em plus 0.5em minus 0.4em\relax IEEE, 2021, pp. 1--8.

\bibitem{chen2022geoconv}
Y.~Chen, G.~Song, Z.~Shao, J.~Cai, T.-J. Cham, and J.~Zheng, ``Geoconv:
  Geodesic guided convolution for facial action unit recognition,''
  \emph{Pattern Recognition}, vol. 122, p. 108355, 2022.

\bibitem{srivastava2014dropout}
N.~Srivastava, G.~Hinton, A.~Krizhevsky, I.~Sutskever, and R.~Salakhutdinov,
  ``Dropout: a simple way to prevent neural networks from overfitting,''
  \emph{The journal of machine learning research}, vol.~15, no.~1, pp.
  1929--1958, 2014.

\bibitem{hou2019weighted}
S.~Hou and Z.~Wang, ``Weighted channel dropout for regularization of deep
  convolutional neural network,'' in \emph{Proceedings of the AAAI Conference
  on Artificial Intelligence}, vol.~33, no.~01, 2019, pp. 8425--8432.

\bibitem{ding2021channel}
Y.~Ding, S.~Dong, Y.~Tong, Z.~Ma, B.~Xiao, and H.~Ling, ``Channel dropblock: An
  improved regularization method for fine-grained visual classification,'' in
  \emph{BMVC}, 2021.

\bibitem{kong2022reflash}
X.~Kong, X.~Liu, J.~Gu, Y.~Qiao, and C.~Dong, ``Reflash dropout in image
  super-resolution,'' in \emph{Proceedings of the IEEE/CVF Conference on
  Computer Vision and Pattern Recognition}, 2022, pp. 6002--6012.

\bibitem{dosovitskiy2020image}
A.~Dosovitskiy, L.~Beyer, A.~Kolesnikov, D.~Weissenborn, X.~Zhai,
  T.~Unterthiner, M.~Dehghani, M.~Minderer, G.~Heigold, S.~Gelly \emph{et~al.},
  ``An image is worth 16x16 words: Transformers for image recognition at
  scale,'' in \emph{International Conference on Learning Representations},
  2020.

\bibitem{yosinski2014transferable}
J.~Yosinski, J.~Clune, Y.~Bengio, and H.~Lipson, ``How transferable are
  features in deep neural networks?'' \emph{Advances in neural information
  processing systems}, vol.~27, 2014.

\bibitem{ganin2015unsupervised}
Y.~Ganin and V.~Lempitsky, ``Unsupervised domain adaptation by
  backpropagation,'' in \emph{International conference on machine
  learning}.\hskip 1em plus 0.5em minus 0.4em\relax PMLR, 2015, pp. 1180--1189.

\bibitem{ge2023algrnet}
X.~Ge, J.~M. Jose, P.~Wang, A.~Iyer, X.~Liu, and H.~Han, ``Algrnet:
  Multi-relational adaptive facial action unit modelling for face
  representation and relevant recognitions,'' \emph{IEEE Transactions on
  Biometrics, Behavior, and Identity Science}, 2023.

\bibitem{shao2021jaa}
Z.~Shao, Z.~Liu, J.~Cai, and L.~Ma, ``Jaa-net: joint facial action unit
  detection and face alignment via adaptive attention,'' \emph{International
  Journal of Computer Vision}, vol. 129, pp. 321--340, 2021.

\bibitem{wang2022action}
L.~Wang, J.~Qi, J.~Cheng, and K.~Suzuki, ``Action unit detection by exploiting
  spatial-temporal and label-wise attention with transformer,'' in
  \emph{Proceedings of the IEEE/CVF Conference on Computer Vision and Pattern
  Recognition}, 2022, pp. 2470--2475.

\bibitem{chang2022knowledge}
Y.~Chang and S.~Wang, ``Knowledge-driven self-supervised representation
  learning for facial action unit recognition,'' in \emph{Proceedings of the
  IEEE/CVF Conference on Computer Vision and Pattern Recognition}, 2022, pp.
  20\,417--20\,426.

\bibitem{li2019self}
Y.~Li, J.~Zeng, S.~Shan, and X.~Chen, ``Self-supervised representation learning
  from videos for facial action unit detection,'' in \emph{CVPR}, 2019, pp.
  10\,924--10\,933.

\bibitem{peng2018weakly}
G.~Peng and S.~Wang, ``Weakly supervised facial action unit recognition through
  adversarial training,'' in \emph{CVPR}, 2018, pp. 2188--2196.

\bibitem{li2020learning}
Y.~Li, J.~Zeng, and S.~Shan, ``Learning representations for facial actions from
  unlabeled videos,'' \emph{IEEE Transactions on Pattern Analysis and Machine
  Intelligence}, 2020.

\bibitem{saito2018maximum}
K.~Saito, K.~Watanabe, Y.~Ushiku, and T.~Harada, ``Maximum classifier
  discrepancy for unsupervised domain adaptation,'' in \emph{Proceedings of the
  IEEE conference on computer vision and pattern recognition}, 2018, pp.
  3723--3732.

\bibitem{lee2021dranet}
S.~Lee, S.~Cho, and S.~Im, ``Dranet: Disentangling representation and
  adaptation networks for unsupervised cross-domain adaptation,'' in
  \emph{Proceedings of the IEEE/CVF conference on computer vision and pattern
  recognition}, 2021, pp. 15\,252--15\,261.

\bibitem{chen2021joint}
H.~Chen, Y.~Wang, B.~Lagadec, A.~Dantcheva, and F.~Bremond, ``Joint generative
  and contrastive learning for unsupervised person re-identification,'' in
  \emph{Proceedings of the IEEE/CVF conference on computer vision and pattern
  recognition}, 2021, pp. 2004--2013.

\bibitem{morerio2017curriculum}
P.~Morerio, J.~Cavazza, R.~Volpi, R.~Vidal, and V.~Murino, ``Curriculum
  dropout,'' in \emph{Proceedings of the IEEE International Conference on
  Computer Vision}, 2017, pp. 3544--3552.

\bibitem{ghiasi2018dropblock}
G.~Ghiasi, T.-Y. Lin, and Q.~V. Le, ``Dropblock: A regularization method for
  convolutional networks,'' \emph{Advances in neural information processing
  systems}, vol.~31, 2018.

\bibitem{lee2019drop}
S.~Lee, D.~Kim, N.~Kim, and S.-G. Jeong, ``Drop to adapt: Learning
  discriminative features for unsupervised domain adaptation,'' in
  \emph{Proceedings of the IEEE/CVF international conference on computer
  vision}, 2019, pp. 91--100.

\bibitem{zeng2020corrdrop}
Y.~Zeng, T.~Dai, and S.-T. Xia, ``Corrdrop: Correlation based dropout for
  convolutional neural networks,'' in \emph{ICASSP 2020-2020 IEEE International
  Conference on Acoustics, Speech and Signal Processing (ICASSP)}.\hskip 1em
  plus 0.5em minus 0.4em\relax IEEE, 2020, pp. 3742--3746.

\bibitem{guo2023domaindrop}
J.~Guo, L.~Qi, and Y.~Shi, ``Domaindrop: Suppressing domain-sensitive channels
  for domain generalization,'' in \emph{Proceedings of the IEEE/CVF
  International Conference on Computer Vision}, 2023, pp. 19\,114--19\,124.

\bibitem{efraimidis2006weighted}
P.~S. Efraimidis and P.~G. Spirakis, ``Weighted random sampling with a
  reservoir,'' \emph{Information processing letters}, vol.~97, no.~5, pp.
  181--185, 2006.

\bibitem{zhang2014bp4d}
X.~Zhang, L.~Yin, J.~F. Cohn, S.~Canavan, M.~Reale, A.~Horowitz, P.~Liu, and
  J.~M. Girard, ``Bp4d-spontaneous: a high-resolution spontaneous 3d dynamic
  facial expression database,'' \emph{Image and Vision Computing}, vol.~32,
  no.~10, pp. 692--706, 2014.

\bibitem{zhang2016multimodal}
Z.~Zhang, J.~M. Girard, Y.~Wu, X.~Zhang, P.~Liu, U.~Ciftci, S.~Canavan,
  M.~Reale, A.~Horowitz, H.~Yang \emph{et~al.}, ``Multimodal spontaneous
  emotion corpus for human behavior analysis,'' in \emph{Proceedings of the
  IEEE conference on computer vision and pattern recognition}, 2016, pp.
  3438--3446.

\bibitem{mavadati2013disfa}
S.~M. Mavadati, M.~H. Mahoor, K.~Bartlett, P.~Trinh, and J.~F. Cohn, ``Disfa: A
  spontaneous facial action intensity database,'' \emph{IEEE Transactions on
  Affective Computing}, vol.~4, no.~2, pp. 151--160, 2013.

\bibitem{girard2017sayette}
J.~M. Girard, W.-S. Chu, L.~A. Jeni, and J.~F. Cohn, ``Sayette group formation
  task (gft) spontaneous facial expression database,'' in \emph{2017 12th IEEE
  international conference on automatic face \& gesture recognition (FG
  2017)}.\hskip 1em plus 0.5em minus 0.4em\relax IEEE, 2017, pp. 581--588.

\bibitem{zou2020joint}
Y.~Zou, X.~Yang, Z.~Yu, B.~V. Kumar, and J.~Kautz, ``Joint disentangling and
  adaptation for cross-domain person re-identification,'' in \emph{ECCV}.\hskip
  1em plus 0.5em minus 0.4em\relax Springer, 2020, pp. 87--104.

\bibitem{selvaraju2017grad}
R.~R. Selvaraju, M.~Cogswell, A.~Das, R.~Vedantam, D.~Parikh, and D.~Batra,
  ``Grad-cam: Visual explanations from deep networks via gradient-based
  localization,'' in \emph{Proceedings of the IEEE international conference on
  computer vision}, 2017, pp. 618--626.

\end{thebibliography}
	
\begin{IEEEbiography}[{\includegraphics[width=1in,height=1.25in,clip,keepaspectratio]{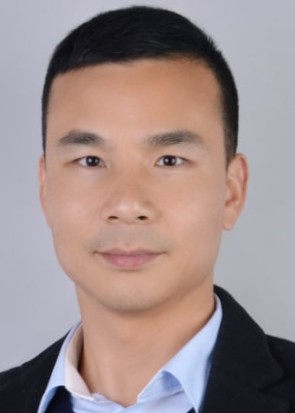}}]
	{Yong Li} is currently an Associate Professor with the School of Computer Science and Engineering, Southeast University, China. He received the Ph.D. degree from Institute of Computing Technology (ICT), Chinese Academy of Sciences in 2020. He has previously held a position at Nanjing University of Science and Technology. In 2023-2024, he worked as a Research Fellow in City Universitiy of Hong Kong and Nanyang Technological University.  His research interests include face-related deep learning, human-centered affective computing. His research results have been expounded in more than 40 publications at prestigious journals and prominent conferences, such as IEEE TPAMI, IEEE TIP, IEEE TAC, IEEE TMM, NeurIPS, CVPR, ICCV.  For more information, please visit his personal website: https://mysee1989.github.io/.
\end{IEEEbiography}

\begin{IEEEbiography}[{\includegraphics[width=1in,height=1.25in,clip,keepaspectratio]{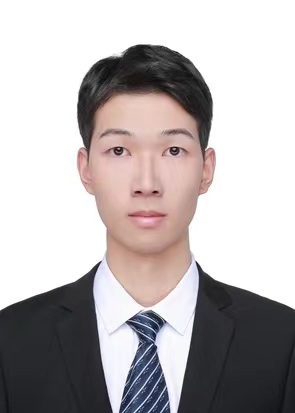}}]
	{Yi Ren} received the bachelor's degree in Software Engineering from the Zhengzhou University, Zhengzhou, China, in 2023. His research interests include deep learning, affective computing, and multimodal learning.
\end{IEEEbiography}

\begin{IEEEbiography}[{\includegraphics[width=1in,height=1.25in,clip,keepaspectratio]{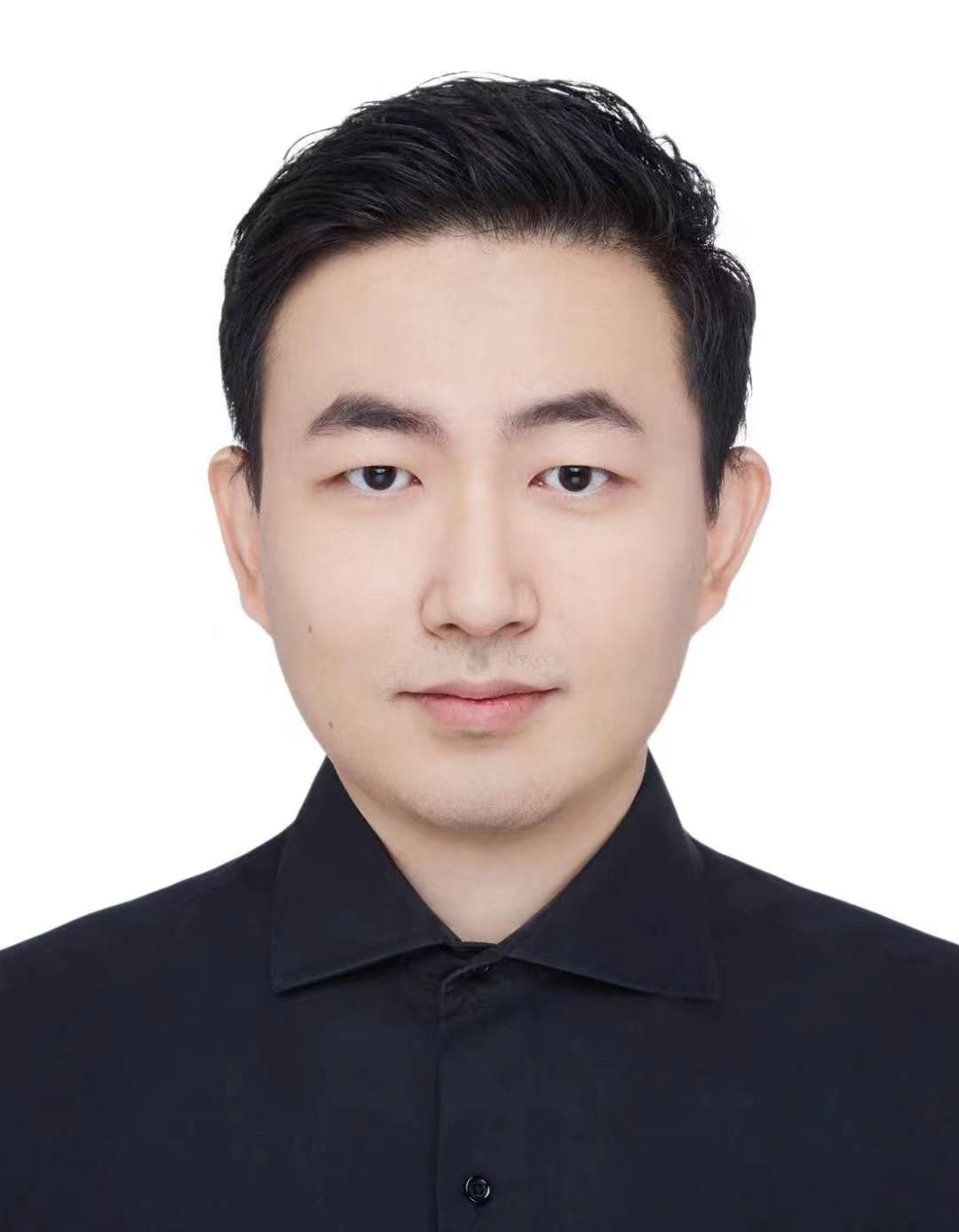}}]
	{Xuesong Niu} received his B.S. degree from Nankai University, and his Ph.D. degree from Institute of Computing Technology (ICT), Chinese Academy of Sciences (CAS), in 2016 and 2021, respectively, both in computer science. Currently, he is working as a researcher engineer in the Beijing Institute for General Artificial Intelligence, working on the problem of human and environment understanding. His research interests include the deep perception and understanding of the real world, especially understanding human face, body and the environment from 2D images and 3D representations.  He has authored or co-authored several papers in refereed journals and conferences including IEEE Trans. IP/CSVT, CVPR, ECCV, NeurIPS. He was a recipient of the IEEE FG 2019 Best Poster Presentation Award.
\end{IEEEbiography}

\begin{IEEEbiography}
	[{\includegraphics[width=1in,height=1.25in,clip,keepaspectratio]{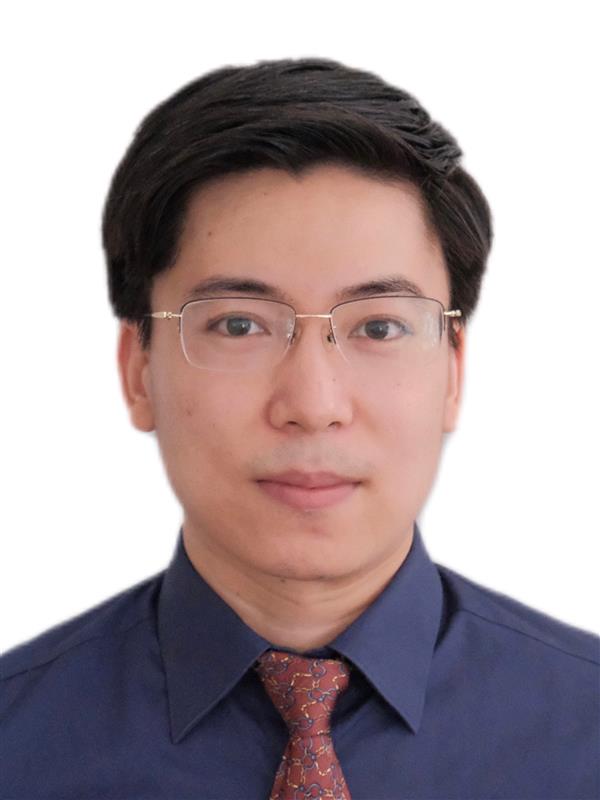}}]{Yi Ding}(Graduate student member, IEEE) earned a Ph.D. in Computer Science and Engineering from Nanyang Technological University, Singapore, in 2023, a master's degree in Electrical and Electronics Engineering from the same university in 2018, and a bachelor's degree in Information Science and Technology from Donghua University, Shanghai, China, in 2017. His research interests encompass brain-computer interface, deep learning, graph neural networks, neural signal processing, affective computing, and multimodal learning.
\end{IEEEbiography}

\begin{IEEEbiography}
	[{\includegraphics[width=1in,height=1.25in,clip,keepaspectratio]{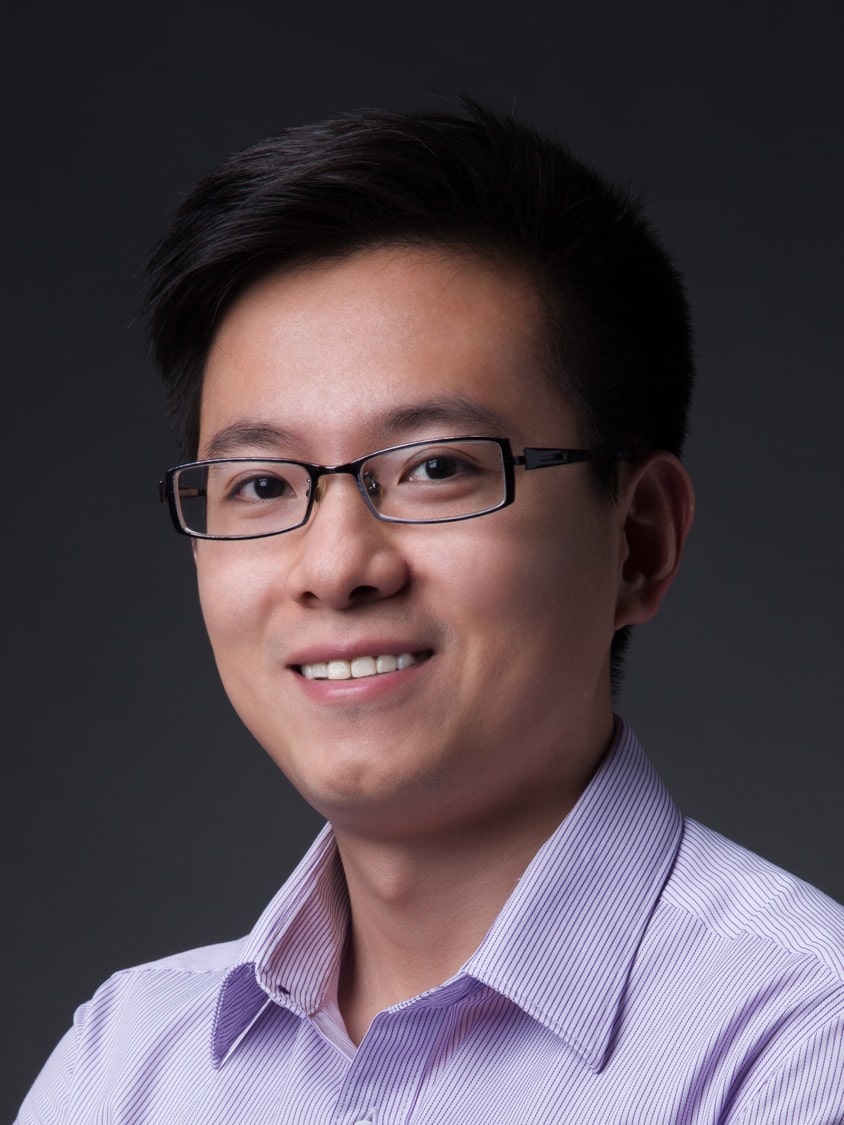}}]{Xiu-Shen Wei}(Member, IEEE) is a professor with the School of Computer Science and Engineering,
	Southeast University, China. He was a program chair
	for the workshops associated with ICCV, IJCAI, ACM
	Multimedia, etc. He has also served as an area chair or
	senior program member with CVPR, AAAI, IJCAI,
	ICME, BMVC, a guest editor of Pattern Recognition
	Journal, and a tutorial chair for Asian Conference on
	Computer Vision (ACCV) 2022.
\end{IEEEbiography}

\begin{IEEEbiography}
	[{\includegraphics[width=1in,height=1.25in,clip,keepaspectratio]{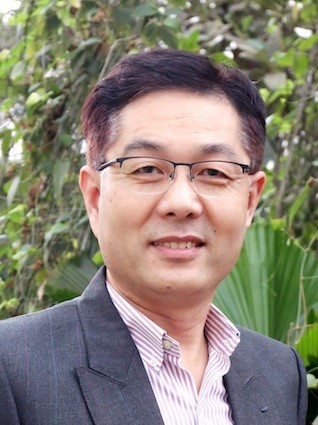}}]{Cuntai Guan}(Fellow, IEEE) received his Ph.D. degree from Southeast University, China, in 1993. He is a President’s Chair Professor in the School of Computer Science and Engineering, Nanyang Technological University, Singapore. He is the Director of the Artificial Intelligence Research Institute, Director of the Centre for Brain-Computing Research, and Co-Director of S-Lab for Advanced Intelligence. His research interests include brain-computer interfaces, machine learning, medical signal and image processing, artificial intelligence, and neural and cognitive rehabilitation. He is a recipient of the Annual BCI Research Award (first prize), King Salman Award for Disability Research, IES Prestigious Engineering Achievement Award, Achiever of the Year (Research) Award, and Finalist of President Technology Award. He is also an elected Fellow of the US National Academy of Inventors (NAI), the Academy of Engineering Singapore (SAEng), and the American Institute for Medical and Biological Engineering (AIMBE).
\end{IEEEbiography}

\end{document}